\documentclass[letterpaper]{article}

\usepackage[preprint]{aaai2027}

\usepackage[hyphens]{url}  %
\usepackage{graphicx} %
\usepackage{natbib}  %
\usepackage{caption} %
\usepackage{algorithm}
\usepackage{algorithmic}

\usepackage{newfloat}
\usepackage{listings}
\DeclareCaptionStyle{ruled}{labelfont=normalfont,labelsep=colon,strut=off} %
\floatstyle{ruled}
\newfloat{listing}{tb}{lst}{}
\floatname{listing}{Listing}
\usepackage{booktabs}
\usepackage{microtype}
\usepackage{subfigure}
\usepackage{graphicx}
\usepackage{multirow}

\usepackage{natbib}

\usepackage{amsmath}
\usepackage{amssymb}
\usepackage{mathtools}
\usepackage{amsthm}

\usepackage{amsfonts}       %
\usepackage{nicefrac}       %
\usepackage{microtype}      %
\usepackage{xcolor}         %
\usepackage{amsbsy}
\usepackage{graphicx}
\usepackage{bm}
\usepackage{svg}
\usepackage{siunitx}

\theoremstyle{plain}

\theoremstyle{definition}

\theoremstyle{remark}

\newcommand{\var}[0]{\text{var}}
\newcommand{\diag}[0]{\text{diag}}
\DeclareMathOperator*{\argmin}{arg\,min}

\usepackage{booktabs}

\definecolor{blue_mpl}{HTML}{1F77B4}
\definecolor{orange_mpl}{HTML}{FF7F0E}

\usepackage[hidelinks=true]{hyperref}

\begin{document}

\title{Tracing sources of epistemic uncertainty in deep learning predictions:\\homo- and hetero-scedastic linearized estimators}

\author{
    Pierre Nodet\equalcontrib\corresponding,\\
    Thomas George\equalcontrib\corresponding
}
\affiliations{
    Orange Research, Châtillon, France\\
    pierre.nodet@orange.com, thomas.george@orange.com
}

\maketitle

\begin{abstract}We adapt two classical statistical estimators for quantifying uncertainty to modern deep learning, in order to provide clearer insights into uncertainty attributable to two sources : aleatoric uncertainty, or locally scarce data. Our approach leverages recent advances in approximate Fisher Information Matrices, to enable scaling to actual architectures. Experimental results demonstrate how each test points is differentially impacted by both sources, highlighting the practical utility of our estimators in improving the robustness of real-world applications.%
\end{abstract}

\section{Introduction}

AI systems deployed in high-stakes real-world applications require accurately assessing uncertainty associated with a given prediction, for example to fall back to human monitoring if too uncertain. In supervised machine learning, this uncertainty arises from the fact that we observe a target function that we want to estimate only through a given finite training sample (the training dataset). If we suppose this process to be noisy (some labels are unreliable), and the function to estimate to be complex enough (requiring generalization to unseen test instances), then the learning problem is misspecified, so there are test instances where we can confidently give a prediction, and others where our estimation is merely unreliable. The key question is then, \emph{how to predict the uncertainty associated to a prediction on a given test instance}.

More formally, we suppose that there is a true underlying model $p\left(x,y\right)$ on an instance space $\mathcal{X}$ and response space $\mathcal{Y}$, that we observe through a finite training dataset $\mathcal{D}_n:=\left\{ \left(\bm{x}_{i},y_{i}\right)\right\} _{1\leq i\leq n}$, and we want to estimate $p\left(y|x_\text{test}\right)$ for future, held-out, test points $x_\text{test}$. Here, uncertainty comes both from the randomness $p\left(y|x\right)$ of the process that we are trying to estimate (termed aleatoric uncertainty), as well as uncertainty resulting from our \emph{finite-sample knowledge} of this process, both contributing to increasing the epistemic uncertainty of the learned model \citep{hullermeier2021aleatoric}.%

\begin{figure}[t]
    \centering
    \includegraphics[width=0.85\linewidth]{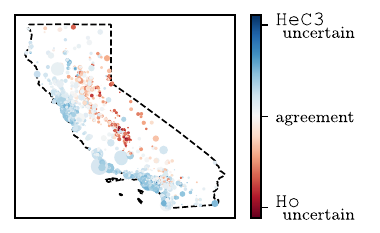}
    \caption{Regression on the California Housing dataset. The circle size corresponds to the prediction error of a MLP. Districts are colored in blue if the model is \texttt{HeC3}-uncertain (regions with higher aleatoric uncertainty) or in red if \texttt{Ho}-uncertain (regions far from training data).}
    \label{fig:housing}
\end{figure}

Focusing on epistemic uncertainty, we seek to identify contributions from 2 different sources to the variance of the prediction for fixed test points: the training dataset is a finite sample of the true generating process, which cannot comprehensively capture its structure in general: instances $\left\{ \bm{x}_{i}\right\} _{1\leq i\leq n}$ only partially cover the instance space, and responses $\left\{ y_{i}\right\} _{1\leq i\leq n}$ are a single realization of a random process. A machine learning algorithm trained on this dataset inherits this limited knowledge, which translates into uncertainty in future test predictions. Ultimately, we would like to be able to answer the question: \emph{how much epistemic uncertainty is there for given test points and what is it attributable to?} In particular, if we want to reduce it, should we preferably \emph{collect more training points in scarce regions} or should we \emph{reduce aleatoric uncertainty} (e.g., by adding informative features, or reducing label noise) ?

Our perspective is inspired by insights from linear regression, which is well-studied with theoretical results in textbooks in statistics, that we apply to deep learning using the (empirical) tangent kernel/features framework \citep{jacot2018neural,fort2020deep,grosse2021taylor}. The originality of our approach is that we not only seek a method to more accurately estimate uncertainty in general, but rather show that the proposed estimators account for different sources of uncertainty. Our \textbf{contributions} are:%

\begin{itemize}
    \item a frequentist perspective on the family of Laplace uncertainty estimators as local solution to a linearized model of the training dynamics of deep networks, allowing us to derive a ridge estimator, and a sandwich estimator in deep learning ;
    \item algorithms \texttt{Ho}-EKFAC and \texttt{HeC3}-EKFAC that enable to scale estimators to actual deep architectures ;
    \item experiments evaluating the proposed estimators, characterizing the different types of uncertainty that they measure on toy models and actual tasks.
\end{itemize}

\subsection{Related works}

This work draws on textbook statistical concepts that have not yet been widely adopted in modern deep learning practices. We argue that this is partly due to the initial misconception that these concepts would be impractical to scale to modern architectures with a large number of parameters. The closest works that we are aware of are \citet{rivals2000construction} and \citet{tibshirani_comparison_1996} who give confidence intervals estimators for non-linear models by linearization. Two main differences are that they were limited to the smaller architectures that were used at that time, and they did not use their tools to explicitly disentangle sources of uncertainty. More recently, delta estimators have also been shown to scale to million-parameter architectures and give useful estimators even using crude diagonal approximations \citep{schmitt_general_2025}, or the Lanczos iterative method for estimating the top spectrum \citep{nilsen_epistemic_2022} of the Fisher Information Matrix, an object central to most variance estimators.

From a practical standpoint, these estimators require computing inverse Hessian or inverse Fisher Information Matrix of deep networks, which can be found in influence functions \citep{koh2017understanding,grosse2023studying}, efficient optimization \citep{amari_natural_1998}, and continual learning \citep{liu_rotate_2018} to cite a few. All of these works face the same scaling challenge for over-parametrized methods that they handle by resorting to approximate methods
\citep{schraudolph_fast_2002, martens2015optimizing, george2018fast, wang2025better}.

Quantifying epistemic uncertainty in deep learning is active research, with popular methods such as ensembles of deep networks trained using different bootstrapped training subsets \citep{breiman1996bagging}, from different initial parameters \citep{lakshminarayanan2017simple}, using Monte-Carlo estimates based on successive randomly dropped-out neurons \citep{gal_dropout_2016}, or inspired by Bayesian methods \citep{ritter_scalable_2018,immer_improving_2021}. In contrast, our approach is based on a single trained model, and it is able to disentangle uncertainty attributable to noise, from uncertainty attributable to the finite size of the training set.

\section{Background}

Let $p\left(x,y\right)$ represent a true unknown process on an instance space $\mathcal{X}$ and response space $\mathcal{Y}$, that we observe through a finite training dataset $\mathcal{D}_n:=\left\{ \left(\bm{x}_{i},y_{i}\right)\right\} _{1\leq i\leq n}$ with independent and identically distributed (IID) samples, and we want to learn a parametric predictor $\hat{f}\left(\bm{x};\hat{\bm{w}}\left(\mathcal{D}_{n}\right)\right)$ in order to estimate the Bayes-optimal predictor $y^*\left(\bm{x}\right)=\mathbb{E}_{y|\bm{x}}\left[y\right]$, where we made explicit that parameters $\hat{\bm{w}}\left(\mathcal{D}_{n}\right)$ are learned from the training dataset $\mathcal{D}_{n}$. We restrict our study to the case of regression: $\mathcal{Y}$ is assimilated to $\mathbb{R}$, and parameters $\hat{\bm{w}}\left(\mathcal{D}_{n}\right)$ are learned by seeking a minimizer of the mean squared error (MSE) with a ridge penalty $\lambda\ge0$:
\begin{align}\hat{\bm{w}}\left(\mathcal{D}_{n}\right)\in\underset{\bm{w}}{\argmin}\sum_{i=1}^{n}\left(\hat{f}\left(\bm{x}_{i};\bm{w}\right)-y_{i}\right)^{2}+\lambda\left\Vert \bm{w}\right\Vert_2 ^{2}\label{eq:min_emp_risk}\end{align}
\paragraph{Uncertainty and variance} Our goal is to quantify the reliability of predictions on future test points $\bm{x}_{\text{te}}$, by obtaining properties on their squared loss to the Bayes predictor $\left(\hat{f}\left(\bm{x}_{\text{te}};\hat{\bm{w}}\left(\mathcal{D}_{n}\right)\right)-y^{*}\left(\bm{x}_{\text{te}}\right)\right)^{2}$. A classical quantity is to compute the expected value of this loss, had we obtained a different training set, now treated as a random variable $D_{n}\sim\ensuremath{p\left(x,y\right)}^{n}$. This quantity naturally decomposes into \emph{bias} and \emph{variance} terms, here written for a \emph{single} prediction:
\begin{gather*}
    \mathbb{E}_{D_{n}}\left[\left(\hat{f}\left(\bm{x}_{\text{te}};\hat{\bm{w}}\left(D_{n}\right)\right)-y^{*}\left(\bm{x}_{\text{te}}\right)\right)^{2}\right]=\\\mathbb{E}_{D_{n}}\underbrace{\left[\left({\color{olive}\hat{f}\left(\bm{x}_{\text{te}};\hat{\bm{w}}\left(D_{n}\right)\right)-\mathbb{E}_{D_{n}}\left[\hat{f}\left(\bm{x}_{\text{te}};\hat{\bm{w}}\left(D_{n}\right)\right)\right]}\right)^{2}\right]}_{\text{variance}}\\+\underbrace{\left({\color{violet}y^{*}\left(\bm{x}_{\text{te}}\right)-\mathbb{E}_{D_{n}}\left[\hat{f}\left(\bm{x}_{\text{te}};\hat{\bm{w}}\left(D_{n}\right)\right)\right]}\right)^{2}}_{\text{bias}}
\end{gather*}

The bias involves the Bayes predictor $y^*$, which is unknown in general. The variance measures fluctuations of the estimator $\hat{f}$ around its expected value, which, in contrast to the bias, involves only quantities calculated using the data. In the remainder, we focus on estimating this variance term.

\subsection{Linear regression: dependence on finite samples and resulting heteroscedasticity}\label{subsec:bg_var_uncertainty}

We start with the simpler setup of linear regression, where our parametric estimator of the prediction function takes the form $f\left(\bm{x};\bm{w}\right)=\bm{x}^{\top}\bm{w}$. The training data $\mathcal{D}_n:=\left(X,\bm{y}\right)$ consists of input space observations $\left\{ \bm{x}_{i}\right\} _{1\leq i\leq n}$ stacked in the $n\times d$ design matrix $X$, and corresponding responses $y_i\sim p\left(y|\bm{x}_i\right)$ grouped in the $n$-vector $\bm{y}$. We also define $\boldsymbol{\varepsilon}$ as the $n$-vector of noise, i.e. the difference between the Bayes predictor, and the observed (potentially noisy) label $\varepsilon_{i}\coloneqq y_{i}-y^{*}\left(x_{i}\right)$. We denote by $\bm{w}^*$ the (ideal) parameters of the best linear model that we would obtain if we minimized the squared error on the population $\bm{w}^{*}\in\underset{\bm{w}}{\argmin}\,\mathbb{E}_{p\left(\bm{x},y\right)}\left[\left(\bm{x}^{\top}\bm{w}-y\right)^{2}\right]$\footnote{Except in degenerate cases, this is uniquely defined since we are here dealing with the population risk.}. Recall that we made no assumption on the true generating process $p\left(\bm{x},y\right)$, in particular it might not be well approximated by its linear estimator, in which case we say that our model (linear) is \emph{misspecified}.

We denote by $\hat{\Sigma}_{\lambda}:=X^{\top}X+\lambda I$ the ridge regularized uncentered sample covariance, and $\hat{\Sigma}:=\hat{\Sigma}_{0}$ the unregularized one. As a classical textbook result of the linear MSE problem, we obtain a closed-form solution to the estimator of the prediction for a test point $f\left(\bm{x}_{\text{te}};\hat{\bm{w}}\left(\mathcal{D}_{n}\right)\right)$, assuming $\lambda>0$ or $\hat{\Sigma}$ is full rank (or both): 
\begin{align}
\hat{y}\left(\bm{x}_{\text{te}},X,\boldsymbol{\varepsilon}\right)=\bm{x}_{\text{te}}^{\top}\hat{\Sigma}_{\lambda}^{-1}X^{\top}\bm{y}\label{eq:OLS_pred}
\end{align}
We write $\bm{y}=y^{*}\left(X\right)-X\bm{w}^{*}+X\bm{w}^{*}+\boldsymbol{\varepsilon}$ to explicitly show the effect of misspecification, denoted $\bm{\delta}(X) := y^*(X)-X\bm{w}^*$. The test prediction further decomposes as:
\begin{multline}\hat{y}\left(\bm{x}_{\text{te}},X,\boldsymbol{\varepsilon}\right)=\bm{x}_{\text{te}}^{\top}\bm{w}^{*}+\underbrace{\bm{x}_{\text{te}}^{\top}\hat{\Sigma}_{\lambda}^{-1}X^{\top}\boldsymbol{\varepsilon}}_{\text{random error}}\\+\underbrace{\bm{x}_{\text{te}}^{\top}\hat{\Sigma}_{\lambda}^{-1}X^{\top}\bm{\delta}(X)}_{\text{misspecification error}}-\underbrace{\lambda\,\bm{x}_{\text{te}}^{\top}\hat{\Sigma}_{\lambda}^{-1}\bm{w}^{*}}_{\text{ridge shrinkage}}
\end{multline}
This form highlights the dependence of the estimator on the particular samples $X$ and $\boldsymbol{\varepsilon}$. Different samples would provide different predictors, and the link between ($X$, $\boldsymbol{\varepsilon}$), and the estimator 
is here made explicit. This also involves the query instance $\bm{x}_{\text{te}}$, and a direct consequence of the choice of linear model as hypothesis class is this particular functional dependence between $\left(X,\boldsymbol{\varepsilon},\bm{x}_{\text{te}}\right)$ and the prediction at a given test point, suggesting that prediction variance will similarly depend on the test instance.

In the following, we will stop explicitly writing the dependency on random variables $X$ and $\boldsymbol{\varepsilon}$ to lighten the notations.

\subsubsection{Fixed design variance} In fixed design, the set of instances $X$ is considered fixed, and the only random quantity is the response noise $\boldsymbol{\varepsilon}$. The variance of the estimated prediction, \begin{equation}\var_{\boldsymbol{\varepsilon}}\left(\hat{y}\left(\bm{x}_{\text{te}}\right)|X\right)=\var_{\boldsymbol{\varepsilon}}\left(\bm{x}_{\text{te}}^{\top}\hat{\Sigma}_{\lambda}^{-1}X^{\top}\boldsymbol{\varepsilon}|X\right)\label{eq:var_given_X}\end{equation} involves an expectation over the random variable $\boldsymbol{\varepsilon}$. It captures the effect of randomness in $y|\bm{x}$ on the variance of a test prediction. In practice, it covers effects like sensor fluctuations or labeling errors.

\subsubsection{Random design variance} In random design, we assume that the design matrix $X$ (and as a consequence, the sample covariance $\hat{\Sigma}_{\lambda}$) are also random. Using the law of total variance, we obtain
\begin{align}\var_{X,\boldsymbol{\varepsilon}}\left(\hat{y}\left(\bm{x}_{\text{te}}\right)\right)= & \;\mathbb{E}_{X}\left(\var_{\boldsymbol{\varepsilon}}\left(\hat{y}\left(\bm{x}_{\text{te}}\right)|X\right)\right)\nonumber \\&+\var_{X}\left(\mathbb{E}_{\boldsymbol{\varepsilon}}\left(\hat{y}\left(\bm{x}_{\text{te}}\right)|X\right)\right)\label{eq:var_total}
\end{align} which additionally involves computing expectations over $X$: had we collected different training examples, we would have obtained a different predictor.

\paragraph{Effect of misspecification and regularization} By construction, $\mathbb{E}_{\boldsymbol{\varepsilon}}\left[\boldsymbol{\varepsilon}\right]=0$, thus the second term of Eq. \ref{eq:var_total} simplifies in: \begin{align}\mathbb{E}_{\boldsymbol{\varepsilon}}\left(\hat{y}\left(\bm{x}_{\text{te}}\right)|X\right)=\bm{x}_{\text{te}}^{\top}\hat{\Sigma}_{\lambda}^{-1}\left(X^{\top}\bm{\delta}\left(X\right)-\lambda\bm{w}^{*}\right)+\bm{x}_{\text{te}}^{\top}\bm{w}^{*}\label{eq:var_misspec}\end{align}

The impact of a random design on the prediction variance therefore arises as a result of attempting to perform linear regression on a misspecified task \citep{buja2019models}, and as a result of ridge regularization. To the contrary, in well-specified, unregularized ($\lambda=0$) problems, Eq. \ref{eq:var_misspec} cancels out and the variance is similar to that of the fixed design.

In the general case, the random design variance thus reads:
\begin{align}\var_{X,\boldsymbol{\varepsilon}}\left(\hat{y}\left(\bm{x}_{\text{te}}\right)\right)
    =  \;\mathbb{E}_{X}\left(\var_{\boldsymbol{\varepsilon}}\left(\bm{x}_{\text{te}}^{\top}\hat{\Sigma}_{\lambda}^{-1}X^{\top}\boldsymbol{\varepsilon}|X\right)\right)\nonumber \\
       +\var_{X}\left(\bm{x}_{\text{te}}^{\top}\hat{\Sigma}_{\lambda}^{-1}\left(X^{\top}\bm{\delta}\left(X\right)-\lambda\bm{w}^{*}\right)\right)
    \label{eq:var_random_design_misspecified}\end{align}

\paragraph{Heteroscedasticity in prediction variance} Interestingly, both variance terms (Eq. \ref{eq:var_given_X} and Eq. \ref{eq:var_random_design_misspecified}) depend on the choice of test point $\bm{x}_{\text{te}}$. For example, even in a setup with homoscedastic noise in the data-generating process ($\varepsilon$ is independent of $\bm{x}$), the modeling choice of linear models as hypothesis class shapes how uncertainty differently affects test points.

\subsection{Variance estimators for linear regression}\label{subsec:estim_linear}

In practice, we do not have access to the true data-generating process. Our goal is thus to provide estimators of both variance terms using only quantities computed on a given sample of $n$ examples.

\paragraph{Fixed design and homoscedastic noise} For fixed $X$, assuming homoscedastic noise with variance $\sigma^2$ , the variance of the prediction admits a closed-form expression:
\begin{equation}
\var_{\boldsymbol{\varepsilon}}\left(\hat{y}\left(\bm{x}_{\text{te}}\right)|X\right)=\sigma^{2}\,\bm{x}_{\text{te}}^{\top}\hat{\Sigma}_{\lambda}^{-1}\hat{\Sigma}\hat{\Sigma}_{\lambda}^{-1}\bm{x}_{\text{te}}
    \label{eq:var_ridge_fixed}
\end{equation}

When the true noise variance $\sigma^2$ is unknown, we use the estimator $\hat{\sigma}^2=\frac{1}{n}\sum_{i=1}^n\left(y_i-\hat{y}\left(\bm{x}_{i}\right)\right)^2$, which gives the uncertainty under the hypothesis of homoscedastic response noise, henceforth designated as \texttt{Ho}.
\begin{equation}
\hat{\var}_{\text{\texttt{Ho}}}\left(\hat{y}\left(\bm{x}_{\text{te}}\right)|X\right)=\\\hat{\sigma}^{2}\,\bm{x}_{\text{te}}^{\top}\hat{\Sigma}_{\lambda}^{-1}\hat{\Sigma}\hat{\Sigma}_{\lambda}^{-1}\bm{x}_{\text{te}}
    \label{eq:estim_mle_ridge}
\end{equation}

\paragraph{Random design and general noise}

In order to provide an estimator for a random design and heteroscedastic noise, without relying on asymptotic assumptions \citep{white1980heteroskedasticity}, we resort to the jackknife estimator \citep{quenouille1956notes, tukey1958bias} which is valid for finite samples and non-parametric (free of any assumptions). It relies on the variance of the leave-one-out predictions (denoted $\hat{y}^{(-i)}(\bm{x}_{\text{te}})$ when example $i$ is left out) around $\hat{y}(\bm{x}_{\text{te}})$.
\begin{equation}
    \hat{\var}_{\text{JK}}\left(\hat{y}(\bm{x}_{\text{te}})\right)=\sum_{i=1}^n\left(\hat{y}(\bm{x}_{\text{te}})-\hat{y}^{(-i)}(\bm{x}_{\text{te}})\right)^2
    \label{eq:def_jk_haty}
\end{equation}

This definition has the advantage of proposing a simple closed-form formula for linear regression using the Sherman-Morrison formula \citep{sherman1950adjustment}:
$$\hat{y}\left(\bm{x}_{\text{te}}\right)-\hat{y}^{\left(-i\right)}\left(\bm{x}_{\text{te}}\right)=\frac{1}{1-h_{ii}}\bm{x}_{\text{te}}^\top\hat{\Sigma}_\lambda^{-1}\bm{x}_{i}\hat{e}_{i}$$

where $h_{ii}:=\bm{x}_i^{\top}\hat{\Sigma}_\lambda^{-1}\bm{x}_i$ is called the \emph{leverage} and $\hat{e}_i=y_i-\hat{y}(\bm{x}_{i})$ is the residual of the model at $(\bm{x}_{i}, y_i)$.

We define $\hat{u}_i:=\frac{\hat{e}_i}{1-h_{ii}}$ as the jackknife residuals. Substituting in Eq. \ref{eq:def_jk_haty}, we get an estimator of the variance of the prediction at a test point $\bm{x}_{\text{te}}$ that stems from randomness of $(X,\boldsymbol{\varepsilon})$, henceforth denoted as \texttt{HeC3} for heteroscedastic consistency \citep[HC3 in][]{mackinnon1985some}:
\begin{equation}
\hat{\var}_{\text{\texttt{HeC3}}}\left(\hat{y}\left(\bm{x}_{\text{te}}\right)\right)=\bm{x}_{\text{te}}^{\top}\hat{\Sigma}_\lambda^{-1}X^{\top}\diag(\hat{\bm{u}}^2)X\hat{\Sigma}_\lambda^{-1}\bm{x}_{\text{te}}
    \label{eq:estim_jack_lin}
\end{equation}

This estimator has the form of a sandwich estimator where the covariance $\hat{\Sigma}_\lambda^{-1}$ is the \emph{bread}, enclosing the middle term $X^{\top}\diag(\hat{\bm{u}}^2)X$ known as the \emph{meat}.

\paragraph{Relation to Laplace approximations} These estimators closely resemble the closed-form Laplace uncertainty estimators \citep{ritter_scalable_2018, daxberger_laplace_2021}, but come from a frequentist perspective. \texttt{Ho} and \texttt{HeC3} are reminiscent of a posterior with covariance matrices that capture more specific effects.

\subsection{Variance estimators in deep learning}\label{subsec:estim_deep}

We now turn to parametrized estimators $f_{\bm{w}}( \bm{x})$, where the functional $\bm{w}\mapsto f_{\bm{w}}$ is non-linear. This setup comprises neural networks, where parameters $\bm{w}$ are weights and biases of all layers, arranged as a $d$-vector. Given a training dataset $\mathcal{D}_n:=\left\{ \left(\bm{x}_{i},y_{i}\right)\right\} _{1\leq i\leq n}$ of IID samples from $p\left(\bm{x},y\right)$, similar to linear regression, we estimate our parameters $\bm{w}$ by minimizing the MSE, but contrary to linear regression, it does not admit a closed-form solution in general, and needs to be solved by iterative algorithms like stochastic gradient descent \citep[SGD][]{robbins1951stochastic}. We suppose that we have access to a minimizer of the ridge regularized MSE as our estimator on parameters, which we denote $\hat{\bm{w}}$. In order to derive estimators of both variance terms Eq. \ref{eq:var_given_X} and \ref{eq:var_total}, we will linearize the functional $\bm{w}\mapsto f_{\bm{w}}$ in $\bm{w}=\hat{\bm{w}}+\bm{\Delta}\bm{w}$:

\begin{equation}f_{\bm{w}}\left(\cdot\right)=f_{\hat{\bm{w}}}\left(\cdot\right)+\bm{\Delta}\bm{w}^{\top}\bm{\phi}_{\hat{\bm{w}}}\left(\cdot\right)+\text{H.O.T.}\label{eq:tangent_features}\end{equation} where $\phi_{\hat{\bm{w}}}\left(\cdot\right)=\frac{\partial f_{\bm{w}}\left(\cdot\right)}{\partial\bm{w}}\Bigr|_{\bm{w}=\hat{\bm{w}}}$ are called the tangent features \citep{jacot2018neural,chizat2019lazy}. During training of a neural network, it has been empirically shown \citep{fort2020deep} that after a short initial phase of representation learning where tangent features rotated and stretched to adapt to the particular task being learned \citep{baratin2021implicit}, training stabilized in linearly connected modes where the linearization in Eq. \ref{eq:tangent_features} essentially captured the actual training dynamics (higher orders vanish). We thus build our estimators by leveraging the analogy with linear models applied on top of tangent features $\phi_{\hat{\bm{w}}}\left(\cdot\right)$, considered fixed in the vicinity of $\hat{\bm{w}}$, used as an anchor. This is equivalent to making the assumption that orders greater than $1$ are negligible when considering different samples $X$ and $\boldsymbol{\varepsilon}$.

\begin{figure*}[t]
    \centering \includegraphics[width=0.8\textwidth]{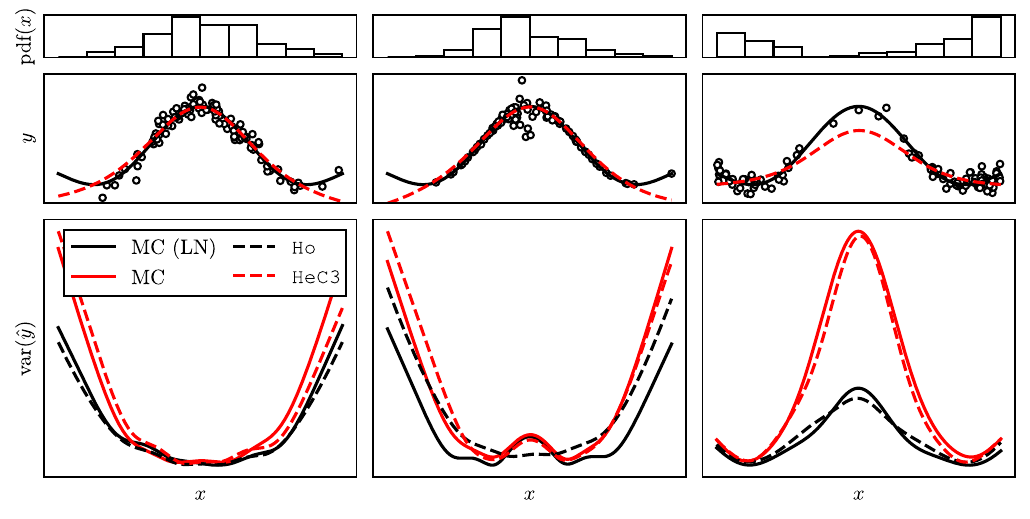}
    \caption{Estimated confidence interval of neural networks with different estimators. Ground truth is computed using Monte Carlo (100 trials), by resampling a full new dataset, or by adding label noise to a fixed dataset.}
    \label{fig:xsinx}
\end{figure*}

For training examples $X$ and corresponding responses $\bm{y}$, we recover the MSE problem with linearized predictor in order to obtain an estimator for $\bm{\Delta}\bm{w}$:
\begin{multline}
    \left\Vert \bm{y}-f_{\bm{w}}\left(X\right)\right\Vert ^{2}+\lambda\left\Vert \bm{w}\right\Vert ^{2}=\\\left\Vert \underbrace{\bm{y}-f_{\hat{\bm{w}}}\left(X\right)}_{:=\text{ pseudo-responses }\bm{r}}-\Phi_{\bm{\hat{w}}}\bm{\Delta}\bm{w}+o\left(\left\Vert \bm{\Delta}\bm{w}\right\Vert \right)\right\Vert ^{2}+\lambda\left\Vert \bm{\hat{w}}+\bm{\Delta w} \right\Vert ^{2}
\end{multline} where $\Phi_{\hat{\bm{w}}}:=\left(\begin{array}{ccc}
            - & \bm{\phi}_{\hat{\bm{w}}}\left(\bm{x}_{1}\right)^{\top} & - \\
              & \vdots                                               \\
            - & \bm{\phi}_{\hat{\bm{w}}}\left(\bm{x}_{n}\right)^{\top} & -
        \end{array}\right)$ are the $n\times d$ stacked tangent features, and $f_{\bm{w}}\left(X\right):=\left(f_{\bm{w}}\left(\bm{x}_{1}\right),\ldots,f_{\bm{w}}\left(\bm{x}_{n}\right)\right)^{\top}$ denotes $f_{\bm{w}}$ applied to every row of $X$. This MSE problem for estimating $\bm{\Delta}\bm{w}$ involves pseudo-responses $\bm{r}$ and transformed design matrix $\Phi_{\bm{w}}$. It admits the closed-form solution: \begin{multline}
\hat{y}\left(\bm{x}_{\text{te}}\right)=f_{\hat{\bm{w}}}\left(\bm{x}_{\text{te}}\right)+\bm{\phi}_{\hat{\bm{w}}}\left(\bm{x}_{\text{te}}\right)^{\top}F_{\bm{\hat{w}}\lambda}^{-1}\left(\Phi_{\bm{\hat{w}}}^{\top}\bm{r}-\lambda\bm{\hat{w}}\right)
\end{multline}
where we defined $F_{\bm{\hat{w}}\lambda}:=\Phi_{\bm{\hat{w}}}^{\top}\Phi_{\bm{\hat{w}}}+\lambda I$ the uncentered covariance of the tangent features. This is similar to Eq. \ref{eq:OLS_pred} in linear regression, allowing directly adapting linear estimators by replacing $X$ by $\Phi_{\bm{w}}$, $\hat{\Sigma}_\lambda$ by $F_{\bm{\hat{w}}\lambda}$, and $\boldsymbol{\varepsilon}$ by $\bm{r}$.

\paragraph{Fixed design and homoscedastic noise in deep learning} We apply the linear estimator of Eq. \ref{eq:estim_mle_ridge}, we obtain:
\begin{multline}
\hat{\text{var}}_{\boldsymbol{\varepsilon}}\left(f_{\hat{\bm{w}}}\left(\bm{x}_{\text{te}}\right)|X\right)=\sigma^{2}\bm{\phi}_{\hat{\bm{w}}}\left(\bm{x}_{\text{te}}\right){}^{\top}F_{\bm{\hat{w}}\lambda}^{-1}F_{\bm{\hat{w}}}F_{\bm{\hat{w}}\lambda}^{-1}\bm{\phi}_{\hat{\bm{w}}}\left(\bm{x}_{\text{te}}\right)\label{eq:var_fix_design_deep}
\end{multline}

and we estimate $\sigma$ with $\hat{\sigma}^2=\frac{1}{n}\sum_{i=1}^n\left(y_i-\hat{y}\left(\bm{x}_{i}\right)\right)^2$ giving the \texttt{Ho} estimator for deep learning:
\begin{multline}
\hat{\text{var}}_{\text{\texttt{Ho}}}\left(f_{\hat{\bm{w}}}\left(\bm{x}_{\text{te}}\right)|X\right)=\hat{\sigma}^{2}\bm{\phi}_{\hat{\bm{w}}}\left(\bm{x}_{\text{te}}\right){}^{\top}F_{\bm{\hat{w}}\lambda}^{-1}F_{\bm{\hat{w}}}F_{\bm{\hat{w}}\lambda}^{-1}\bm{\phi}_{\hat{\bm{w}}}\left(\bm{x}_{\text{te}}\right)\label{eq:estim_mle_deep}
\end{multline}

\paragraph{Random design and general noise in deep learning} Similarly, we directly apply the \texttt{HeC3} estimator derived in linear regression (Eq. \ref{eq:estim_jack_lin}):
\begin{multline}
\hat{\text{var}}_{\text{\texttt{HeC3}}}\left(f_{\hat{\bm{w}}}\left(\bm{x}_{\text{te}}\right)\right)=\\\bm{\phi}_{\hat{\bm{w}}}\left(\bm{x}_{\text{te}}\right){}^{\top}F_{\bm{\hat{w}}\lambda}^{-1}\Phi_{\hat{\bm{w}}}^{\top}\text{diag}\left(\hat{\bm{u}}^{2}\right)\Phi_{\hat{\bm{w}}}F_{\bm{\hat{w}}\lambda}^{-1}\bm{\phi}_{\hat{\bm{w}}}\left(\bm{x}_{\text{te}}\right)\label{eq:estim_jack_deep}
\end{multline}
where $h_{ii}=\bm{\phi}_{\hat{\bm{w}}}\left(\bm{x}_{i}\right){}^{\top}F_{\bm{\hat{w}}\lambda}^{-1}\bm{\phi}_{\hat{\bm{w}}}\left(\bm{x}_{i}\right)$ plays a similar role as the leverage in linear regression, $\hat{e}_i=y_i-\hat{y}(\bm{x}_{i})$ is the residual at $(\bm{x}_i, y_i)$, and $\hat{u}_i=\hat{e}_i/(1-h_{ii})$.

Eq. \ref{eq:estim_mle_deep} and Eq. \ref{eq:estim_jack_deep} serve at our estimators of uncertainty that stems from response noise $\boldsymbol{\varepsilon}$ and finite sampling of $(X,\boldsymbol{\varepsilon})$, respectively, which we now apply in non-linear experiments.

\paragraph{Note on linearization} The use of tangent features is not the only way to linearize deep networks \citep{misiakiewicz2023six}. Another reasonable alternative would be to use latent representations computed at the last layer, and only consider parameters of the last layer, practically considering a linear model on top of these representations.

\begin{figure*}[t]
    \centering
    \includegraphics[width=.74\linewidth]{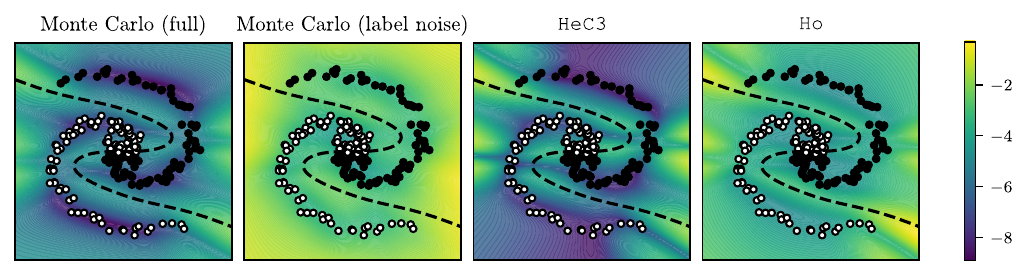}   \includegraphics[width=.25\linewidth]{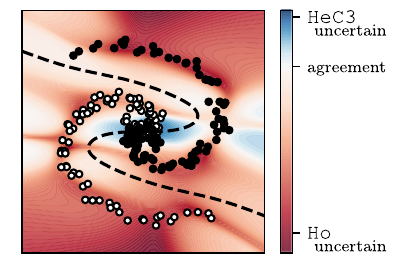}
    \caption{\textbf{(left)} Log-variance estimated using \texttt{Ho} and \texttt{HeC3}. Ground truth variances are estimated using Monte Carlo (100 trials), by resampling a full new dataset or by adding homoscedastic label noise on a fixed dataset. \textbf{(right)} Disagreements between the log-variance computed using \texttt{Ho} and \texttt{HeC3}. Regions of the feature space are colored in blue if the model is uncertain because of the sampling of $(X,\boldsymbol{\varepsilon})$ or in red because of $\boldsymbol{\varepsilon}$.}
    \label{fig:two-moons-comparison}
\end{figure*}

\section{Illustrative Experiments}\label{sec:exp_toy}

\subsection{1D nonlinear regression}

We first conduct an illustrative experiment on a toy 1D nonlinear regression task $x\mapsto\frac{\sin(x)}{x}$ to highlight two scenarios in which the \texttt{Ho} estimator theoretically breaks down relative to the \texttt{HeC3} estimator: heteroscedasticity and the presence of rare training samples, using 3 distinct scenarios:
\begin{enumerate}
    \item normally distributed data, homoscedastic response noise ;
    \item normally distributed data, heteroscedastic response noise concentrated around $x=0$ ;
    \item non-normally distributed data (scarce samples near $x=0$), homoscedastic response noise.
\end{enumerate}

As the dataset is synthetic, we estimate the variance using Monte Carlo sampling \cite{metropolis1949monte}. We employ two procedures: one in which we fully resample the dataset $(X, \bm{y})$ to estimate the total variance, and the other in which the samples $X$ are fixed, and we randomly change the labels for some examples to estimate the variance due to response noise.

Figure \ref{fig:xsinx} shows the comparisons. In the first scenario, where the \texttt{Ho} estimator's assumptions hold, both estimators agree well with their respective Monte Carlo estimates. In the second scenario with heteroscedasticity, as expected, the \texttt{Ho} estimator underestimates the variance when the response noise is higher than the average and overestimates elsewhere, whereas the \texttt{HeC3} estimator closely matches its Monte Carlo estimator. For the third scenario with rare samples, again, both estimators agree well with their respective Monte Carlo estimates, but we can clearly see that the \texttt{Ho} estimator (and also the label-noise Monte Carlo) does not capture the total uncertainty of the model, as it severely underestimates the total uncertainty in the regions where samples are rarer.

\subsection{2D nonlinear classification}\label{sec:exp_toy_two_moons}

We conduct illustrative experiments on a toy nonlinear binary classification task composed of two spirals that slightly overlap near the center of the dataset. 

These experiments are meant to see how both uncertainty estimators behave in two different cases of interest:
\begin{itemize}
    \item On class-overlapping regions (near the center of the dataset), where uncertainty is high because of the intrinsic difficulty of the dataset.
    \item On data-sparse regions (on the tail of the spirals), where uncertainty is high because of the lack of training data.
\end{itemize}

Figure \ref{fig:two-moons-comparison} (right) emphasizes the regions where \texttt{Ho} and \texttt{HeC3} estimators most disagree: First, in class overlapping regions, in blue, where the model locally more frequently makes mistakes on training examples (high jackknife residuals $\hat{\bm{u}}$). Second, they disagree in the tails of the spirals near training datapoints, in red, where the model is sensitive to the label of the few and rare examples.

\subsection{Experiment: California housing dataset}\label{sec:exp_housing}

On the \emph{california-housing} regression dataset, we train a multi-layer MLP with ReLU activations with Adam \citep{kingma2014adam}.

We compare the rankings of the most uncertain test samples from \texttt{Ho} and \texttt{HeC3} in Figure \ref{fig:housing}. The model is less certain about its predictions (\texttt{Ho}-uncertainty) in regions with few districts or in regions with only poor districts, which are underrepresented in training data. On the contrary, in dense regions, but with districts with high value disparities, the model uncertainty is attributable to the intrinsic difficulty of the task (\texttt{HeC3}-uncertainty).

\section{Scaling to deep networks}

Until this point, we have avoided addressing practical considerations, especially questioning whether the estimators we derived theoretically would scale to modern architectures.

The main bottleneck in computing the variance estimators lies in (inverse) matrix-vector products in high dimensions. Indeed, bread and meat matrices of the estimators are $d\times d$ square matrices, $d$ being the parameter count which can scale to $10^{10}$ in recent architectures. These objects are computationally intensive and impossible to store entirely in memory.

\subsection{Eigenvalue-corrected Kronecker Factorization (EKFAC)}
\label{subsec:ekfac}

We resort to using the EKFAC \citep{george2018fast} sparse block approximation of the matrices. We approximate the Fisher information Matrix $\Phi_{\hat{\bm{w}}}^\top \Phi_{\hat{\bm{w}}}$ by a block-diagonal matrix, where each block corresponds to a single layer $(l)$. This block-diagonal approximation ignores cross-layer interactions and only stores, for each layer $(l)$, the block $\Phi_{\hat{\bm{ w}}}^{\top}\Phi_{\hat{\bm {w}}}^{(l)}$. Each block is then approximated with a Kronecker Factorization (KFAC) \citep{martens2015optimizing,grosse2016kronecker}: $\Phi_{\hat{\bm{w}}}^{\top}\Phi_{\hat{\bm{ w}}}^{(l)} \approx A \otimes B$ where $A$ and $B$ are empirical covariance matrices of layer activations and backpropagated gradients, respectively. This Kronecker structure is computationally convenient, since $\left(A \otimes B\right)^{-1} = A^{-1} \otimes B^{-1}$, which makes inverse Hessian vector products (iHVPs) efficient.

EKFAC \citep{george2018fast} additionally improves over KFAC by adjusting the eigenvalues of the Kronecker approximation in its eigenbasis. We first perform the eigendecompositions of $A=U_A\diag(\bm{s}_A)U_A^{\top}$ and $B=U_B\diag(\bm{s}_B)U_B^{\top}$ and define $U = U_A \otimes U_B$ as the Kronecker Factored Eigenbasis (KFE). While KFAC implicitly approximates the eigenvalues of the diagonal block as $\bm{s}_A \otimes \bm{s}_B$, EKFAC used corrected eigenvalues $\bm{s}^*$ chosen to minimize the Frobenius norm error relative to the true diagonal block: $$\bm{s}^{*}=\argmin_{\bm{s}}\Vert\Phi^{\top}_{\hat{\bm{w}}}\Phi_{\hat{\bm{w}}}^{(l)}- U \diag(\bm{s})U^{\top}\Vert_F$$

\subsection{EKFAC for \texttt{HeC3}}

The \texttt{HeC3} estimator in Section \ref{subsec:estim_linear},
\begin{equation*}
\bm{\phi}_{\hat{\bm{w}}}\left(\bm{x}_{\text{te}}\right){}^{\top}F_{\bm{\hat{w}}\lambda}^{-1}\Phi_{\hat{\bm{w}}}^{\top}\text{diag}\left(\hat{\bm{u}}^{2}\right)\Phi_{\hat{\bm{w}}}F_{\bm{\hat{w}}\lambda}^{-1}\bm{\phi}_{\hat{\bm{w}}}\left(\bm{x}_{\text{te}}\right)
\end{equation*}
involves two $d\times d$ matrices: $F_{\bm{\hat{w}}\lambda}$ and $\Phi^{\top}_{\hat{\bm{w}}}\diag(\hat{\bm{u}}^2)\Phi_{\hat{\bm{w}}}$ that we both approximate as EKFAC matrices.

Appendix \ref{app:ekfac} shows that we can further simplify the estimators as the norm of the tangent features projected in the rescaled KFE of both FIMs allowing for efficient evaluation. In a real-world scenario where we seek to estimate the uncertainty of a deployed model with the \texttt{HeC3} and \texttt{Ho} estimators, we can additionally cache the two EKFAC matrices, avoiding the most computationally demanding step (constructing the matrices), and efficiently computing uncertainty estimators.

\section{Experiments: deep networks}
\label{sec:exp_deep}

Our original motivation was to be able to answer: Do we need to \emph{collect more training points in scarce regions} or should we \emph{reduce aleatoric uncertainty} (e.g., by adding informative features, or reducing label noise) to most effectively reduce epistemic uncertainty for given test points?

In control experiments on the MNIST \citep{LeCun2005TheMD} and Fashion-MNIST \citep{xiao2017fashion} datasets, we create random binary meta-classes for which we can control subpopulations: for example, on Fashion-MNIST, one of the generated meta-tasks would be to distinguish between top and bottom garments, rather than the original fine-grained garment recognition task. We create subpopulations of the original data, for which we either remove an original class from the training dataset (rendering the data locally scarce) or corrupt their labels (locally increasing aleatoric uncertainty): for example, in one of the generated runs for Fashion-MNIST, all trousers are randomly labeled, and the bags are missing from the training data. We expect a model trained on this data to be uncertain on test points from noisy or missing subpopulations, but for different reasons. For the noisy subpopulation, the main source of uncertainty should come from heteroscedastic label noise in the training data, captured by the \texttt{HeC3} estimator. By contrast, for missing data, the main source of uncertainty comes as a result of missing representative examples in the training data, captured by both estimators. Comparing estimators thus informs us about the source of uncertainty.

\begin{figure}
    \centering
    \includegraphics[width=0.5\linewidth]{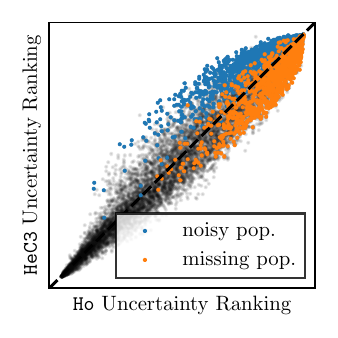}
    \caption{We rank test examples on MNIST using the \texttt{Ho} estimator ($x$-axis) and the \texttt{HeC3} estimator ($y$-axis) of the uncertainty of the neural network predictions. We highlight the missing subpopulation (in orange \textcolor{orange_mpl}{$\bullet$}) and the noisy subpopulation (in blue \textcolor{blue_mpl}{$\bullet$}). Even though both estimators are correlated, a clear pattern emerges to distinguish the two subpopulations.}
    \label{fig:noisy-missing-mnist}
\end{figure}

\begin{table}
\centering
\caption{Corrupted subpopulation detection AUROC, that indicates whether the estimators identify test points from the noisy or missing subpopulations. Average over 5 runs.}
\label{tab:detection-auroc}
\begin{tabular}{llccc}
\toprule
\textbf{Dataset} & \textbf{Pop.} & \textbf{\texttt{Ho}} & \textbf{\texttt{HeC3}} & \textbf{\texttt{HeC3}/\texttt{Ho}} \\
\midrule
\multirow{2}{*}{F-MNIST} & Noisy & 0.67{\tiny$\pm$0.25} & 0.80{\tiny$\pm$0.15} & 0.99{\tiny$\pm$0.01} \\
 & Miss & 0.69{\tiny$\pm$0.19} & 0.68{\tiny$\pm$0.22} & 0.58{\tiny$\pm$0.23} \\
\midrule
\multirow{2}{*}{MNIST} & Noisy & 0.84{\tiny$\pm$0.03} & 0.92{\tiny$\pm$0.02} & 0.97{\tiny$\pm$0.01} \\
 & Miss & 0.80{\tiny$\pm$0.10} & 0.79{\tiny$\pm$0.07} & 0.69{\tiny$\pm$0.09} \\
\bottomrule
\end{tabular}
\end{table}

Table \ref{tab:detection-auroc} shows that the \texttt{Ho} and \texttt{HeC3} estimators capture different sources of uncertainty as their behavior differs on the two subpopulations of interest. The \texttt{HeC3} variance is significantly higher for the noisy subpopulation than for the rest of the dataset, capturing the local task difficulty, whereas the \texttt{Ho} variance is indeed blind to the heteroscedastic noise. We can see that by computing the ratio of the \texttt{HeC3} variance to the \texttt{Ho} variance, we can almost perfectly identify the noisy subpopulation. For the missing subpopulation, both variances are equally high. Figure \ref{fig:noisy-missing-mnist} shows this behavior quantitatively in one of the runs on the MNIST dataset.

We also conducted further experiments to assess whether the \texttt{HeC3} and \texttt{Ho} variances predicted improvements in the model when training on cleaned training examples or with additional training data. For that, we continue training the neural network on the original training data, either with the still-missing subpopulation (to look at improvements when cleaning the training data) or with the noisy subpopulation (to look at improvements when adding training data). We consider that a model has improved on a test point if it goes from wrong to correct prediction. Table \ref{tab:retraining-auroc} shows that, to improve the model on test points with high \texttt{HeC3} variance and low \texttt{Ho} variance, we should clean the training data, and to improve the model on test points with high \texttt{Ho} and \texttt{HeC3} variance, we should add more training data.

\begin{table}
\centering
\caption{Retraining improvement prediction AUROC. Measures whether the uncertainty ranking identifies test points whose predictions improve after cleaning noisy data or adding missing data and then retraining the model. Averaged over 5 runs.}
\label{tab:retraining-auroc}
\begin{tabular}{llccc}
\toprule
\textbf{Dataset} & \textbf{Int.} & \textbf{\texttt{Ho}} & \textbf{\texttt{HeC3}} & \textbf{\texttt{Ho}/\texttt{HeC3}} \\
\midrule
\multirow{2}{*}{F-MNIST} & Clean & 0.70{\tiny$\pm$0.19} & 0.80{\tiny$\pm$0.12} & 0.94{\tiny$\pm$0.03} \\
 & Add & 0.63{\tiny$\pm$0.16} & 0.68{\tiny$\pm$0.16} & 0.70{\tiny$\pm$0.15} \\
\midrule
\multirow{2}{*}{MNIST} & Clean & 0.83{\tiny$\pm$0.02} & 0.89{\tiny$\pm$0.02} & 0.91{\tiny$\pm$0.03} \\
 & Add & 0.80{\tiny$\pm$0.08} & 0.81{\tiny$\pm$0.06} & 0.73{\tiny$\pm$0.07} \\
\bottomrule
\end{tabular}
\end{table}

\paragraph{Uncertainty quantification} We train a ResNet \citep{he2016deep} and a ViT \citep{wang2021not} on the more realistic CIFAR-10 classification task. We compare the \texttt{Ho} and \texttt{HeC3} estimators against a Deep Ensemble \citep{lakshminarayanan2017simple} of 10 models. We rank test examples by their estimated uncertainty and measure how well this ranking predicts whether an example is well-classified by the model, akin to evaluating top-level calibration \citep{perezlebel2023beyond}. We also report the performance of the \texttt{Ho} and \texttt{HeC3} estimators computed only from the last layer, avoiding recourse to EKFAC approximations. For a fraction of the cost of Deep Ensemble (which takes hours for a ViT on CIFAR-10, and much more for deeper models and larger datasets), \texttt{Ho} and \texttt{HeC3} are quite competitive as uncertainty quantifiers (which takes minutes to compute, and rely on techniques shown to scale to LLMs \citep{grosse2023studying}). We also found that \texttt{HeC3} is more robust than \texttt{Ho} when computed solely on the last layer.

\begin{table}[t]
\centering
\caption{Misclassification detection AUROC on test set. Measures whether the uncertainty ranking identifies test points incorrectly predicted. Average over 5 runs.}
\label{tab:misclassification-auroc}
\begin{tabular}{lccc}
\toprule
 & \multicolumn{2}{c}{\textbf{CIFAR10}}  \\
\cmidrule(lr){2-3}
 & ResNet & ViT \\
\midrule
Deep Ensemble (10 models) & 0.88{\tiny$\pm$0.00} & 0.80{\tiny$\pm$0.01} \\
$\texttt{Ho}$-EKFAC (all layers) & 0.85{\tiny$\pm$0.00} & 0.75{\tiny$\pm$0.01} \\
$\texttt{HeC3}$-EKFAC (all layers) & 0.86{\tiny$\pm$0.00} & 0.75{\tiny$\pm$0.01} \\
$\texttt{Ho}$-Dense (last layer) & 0.67{\tiny$\pm$0.03} & 0.68{\tiny$\pm$0.02} \\
$\texttt{HeC3}$-Dense (last layer) & 0.82{\tiny$\pm$0.01} & 0.71{\tiny$\pm$0.02} \\
\bottomrule
\end{tabular}
\end{table}

In the supplementary materials (Appendix \ref{app:xp}), we provide an analysis of the impact of the choice of the regularization parameter $\lambda$ on \texttt{Ho} and \texttt{HeC3} in the supplementary materials (Appendix \ref{app:xp}, Figure \ref{fig:ablation-lambda-sweep}). On small-scale datasets (MNIST and Fashion-MNIST), choosing a value close to the weight decay value used during the optimization was a good heuristic.

Finally, in the supplementary materials (Appendix \ref{app:qual-xp}, we provide qualitative results on a range of neural network architectures, from MLPs to BERT models, on image classification datasets such as CIFAR-10 and on the IMDB \citep{maas2011learning} dataset for sentiment analysis.

\section{Conclusion}

In this work, we highlighted the effect of two different sources of randomness on the uncertainty of machine learning models trained from finite samples, by adapting linear regression estimators to a nonlinear setup (deep learning). These estimators quantify how uncertainty from these 2 different sources differently affects each test point $\bm{x}_{\text{te}}$ of the instance space, which is a direct consequence of the modeling choice of hypothesis class (linear models, or the deep learning architecture).

Traditionally, statistical approaches focus on estimators of model parameters, but the approach using linearization applied to non-linear models only provides estimates of parameter deltas, which are not directly interpretable. Our approach of uncertainty quantification in deep learning is instead to directly estimate the uncertainty in predictions on test points. Furthermore, in structured data, each input dimension typically carries a specific, interpretable meaning, whereas deep learning models excel at handling unstructured data. Individual parameters are not directly associated with input features as opposed to e.g. linear models or regression trees. This distinction underscores the importance of developing estimators of the predictions, not of the parameters.

This work calls for further study along the following lines: First, we aim at more carefully quantifying the practical inaccuracies that result from the assumptions made along the way, and in particular, we should more carefully estimate the quality of linearized estimators, as well as the effect of using approximate inverses. Second, we would like to further break down the variance due to sampling, in order to isolate the effect of $X$ and that of the noise, whereas here the \texttt{HeC3} estimator accounts for both effects simultaneously (both the instance and the response are left out in jackknife estimates). Finally, we plan to derive equivalent estimators in classification settings by adapting logistic regression diagnostics \citep{pregibon1981logistic}, which would notably cover large generative transformers.

\newpage

\bibliography{mybibfile}

@article{white1980heteroskedasticity,
  title     = {A heteroskedasticity-consistent covariance matrix estimator and a direct test for heteroskedasticity},
  author    = {White, Halbert},
  journal   = {Econometrica: journal of the Econometric Society},
  pages     = {817--838},
  year      = {1980},
  publisher = {JSTOR}
}

@article{buja2019models,
  title     = {Models as approximations I},
  author    = {Buja, Andreas and Brown, Lawrence and Berk, Richard and George, Edward and Pitkin, Emil and Traskin, Mikhail and Zhang, Kai and Zhao, Linda},
  journal   = {Statistical Science},
  volume    = {34},
  number    = {4},
  pages     = {523--544},
  year      = {2019},
  publisher = {JSTOR}
}

@inproceedings{he2016deep,
  title     = {Deep residual learning for image recognition},
  author    = {He, Kaiming and Zhang, Xiangyu and Ren, Shaoqing and Sun, Jian},
  booktitle = {Proceedings of the IEEE conference on computer vision and pattern recognition},
  pages     = {770--778},
  year      = {2016}
}

@article{krizhevsky2009learning,
  title     = {Learning multiple layers of features from tiny images},
  author    = {Krizhevsky, Alex and Hinton, Geoffrey and others},
  year      = {2009},
  publisher = {Toronto, ON, Canada}
}

@article{xiao2017fashion,
  title   = {Fashion-mnist: a novel image dataset for benchmarking machine learning algorithms},
  author  = {Xiao, Han and Rasul, Kashif and Vollgraf, Roland},
  journal = {arXiv preprint arXiv:1708.07747},
  year    = {2017}
}

@inproceedings{LeCun2005TheMD,
  title  = {The mnist database of handwritten digits},
  author = {Yann LeCun and Corinna Cortes},
  year   = {2005},
  url    = {https://api.semanticscholar.org/CorpusID:60282629}
}

@article{pace1997sparse,
  title     = {Sparse spatial autoregressions},
  author    = {Pace, R Kelley and Barry, Ronald},
  journal   = {Statistics \& Probability Letters},
  volume    = {33},
  number    = {3},
  pages     = {291--297},
  year      = {1997},
  publisher = {Elsevier}
}

@article{moro2014data,
  title     = {A data-driven approach to predict the success of bank telemarketing},
  author    = {Moro, S{\'e}rgio and Cortez, Paulo and Rita, Paulo},
  journal   = {Decision Support Systems},
  volume    = {62},
  pages     = {22--31},
  year      = {2014},
  publisher = {Elsevier}
}

@article{misiakiewicz2023six,
  title   = {Six lectures on linearized neural networks},
  author  = {Misiakiewicz, Theodor and Montanari, Andrea},
  journal = {arXiv preprint arXiv:2308.13431},
  year    = {2023}
}

@article{wang2021not,
  title   = {Not all images are worth 16x16 words: Dynamic transformers for efficient image recognition},
  author  = {Wang, Yulin and Huang, Rui and Song, Shiji and Huang, Zeyi and Huang, Gao},
  journal = {Advances in neural information processing systems},
  volume  = {34},
  pages   = {11960--11973},
  year    = {2021}
}

@inproceedings{devlin2019bert,
  title     = {Bert: Pre-training of deep bidirectional transformers for language understanding},
  author    = {Devlin, Jacob and Chang, Ming-Wei and Lee, Kenton and Toutanova, Kristina},
  booktitle = {Proceedings of the 2019 conference of the North American chapter of the association for computational linguistics: human language technologies, volume 1 (long and short papers)},
  pages     = {4171--4186},
  year      = {2019}
}

@book{jaeckel1972infinitesimal,
  title     = {The infinitesimal jackknife},
  author    = {Jaeckel, Louis A},
  year      = {1972},
  publisher = {Bell Telephone Laboratories}
}

@article{jacot2018neural,
  title   = {Neural tangent kernel: Convergence and generalization in neural networks},
  author  = {Jacot, Arthur and Gabriel, Franck and Hongler, Cl{\'e}ment},
  journal = {Advances in neural information processing systems},
  volume  = {31},
  year    = {2018}
}

@article{george2018fast,
  title   = {Fast approximate natural gradient descent in a kronecker factored eigenbasis},
  author  = {George, Thomas and Laurent, C{\'e}sar and Bouthillier, Xavier and Ballas, Nicolas and Vincent, Pascal},
  journal = {Advances in neural information processing systems},
  volume  = {31},
  year    = {2018}
}

@inproceedings{martens2015optimizing,
  title        = {Optimizing neural networks with kronecker-factored approximate curvature},
  author       = {Martens, James and Grosse, Roger},
  booktitle    = {International conference on machine learning},
  pages        = {2408--2417},
  year         = {2015},
  organization = {PMLR}
}

@article{liu1989limited,
  title     = {On the limited memory BFGS method for large scale optimization},
  author    = {Liu, Dong C and Nocedal, Jorge},
  journal   = {Mathematical programming},
  volume    = {45},
  number    = {1},
  pages     = {503--528},
  year      = {1989},
  publisher = {Springer}
}

@article{nocedal1980updating,
  title   = {Updating quasi-Newton matrices with limited storage},
  author  = {Nocedal, Jorge},
  journal = {Mathematics of computation},
  volume  = {35},
  number  = {151},
  pages   = {773--782},
  year    = {1980}
}

@article{grosse2023studying,
  title   = {Studying large language model generalization with influence functions},
  author  = {Grosse, Roger and Bae, Juhan and Anil, Cem and Elhage, Nelson and Tamkin, Alex and Tajdini, Amirhossein and Steiner, Benoit and Li, Dustin and Durmus, Esin and Perez, Ethan and others},
  journal = {arXiv preprint arXiv:2308.03296},
  year    = {2023}
}

@article{quenouille1956notes,
  title     = {Notes on bias in estimation},
  author    = {Quenouille, Maurice H},
  journal   = {Biometrika},
  volume    = {43},
  number    = {3/4},
  pages     = {353--360},
  year      = {1956},
  publisher = {JSTOR}
}

@article{chizat2019lazy,
  title   = {On lazy training in differentiable programming},
  author  = {Chizat, Lenaic and Oyallon, Edouard and Bach, Francis},
  journal = {Advances in neural information processing systems},
  volume  = {32},
  year    = {2019}
}

@article{fort2020deep,
  title   = {Deep learning versus kernel learning: an empirical study of loss landscape geometry and the time evolution of the neural tangent kernel},
  author  = {Fort, Stanislav and Dziugaite, Gintare Karolina and Paul, Mansheej and Kharaghani, Sepideh and Roy, Daniel M and Ganguli, Surya},
  journal = {Advances in Neural Information Processing Systems},
  volume  = {33},
  pages   = {5850--5861},
  year    = {2020}
}

@article{cox1995asymptotic,
  title     = {Asymptotic confidence bands for generalized nonlinear regression models},
  author    = {Cox, Christopher and Ma, Guangqin},
  journal   = {Biometrics},
  pages     = {142--150},
  year      = {1995},
  publisher = {JSTOR}
}

@inproceedings{baratin2021implicit,
  title        = {Implicit regularization via neural feature alignment},
  author       = {Baratin, Aristide and George, Thomas and Laurent, C{\'e}sar and Hjelm, R Devon and Lajoie, Guillaume and Vincent, Pascal and Lacoste-Julien, Simon},
  booktitle    = {International Conference on Artificial Intelligence and Statistics},
  pages        = {2269--2277},
  year         = {2021},
  organization = {PMLR}
}

@article{tukey1958bias,
  author  = {Tukey, John W.},
  title   = {Bias and Confidence in Not-quite Large Samples (Preliminary Report)},
  journal = {Annals of Mathematical Statistics},
  volume  = {29},
  number  = {2},
  pages   = {614},
  year    = {1958},
  jstor   = {2237363}
}

@article{schraudolph_fast_2002,
  title    = {Fast {Curvature} {Matrix}-{Vector} {Products} for {Second}-{Order} {Gradient} {Descent}},
  volume   = {14},
  issn     = {0899-7667},
  url      = {https://ieeexplore.ieee.org/abstract/document/6788902},
  doi      = {10.1162/08997660260028683},
  number   = {7},
  urldate  = {2025-10-06},
  journal  = {Neural Computation},
  author   = {Schraudolph, Nicol N.},
  month    = jul,
  year     = {2002},
  pages    = {1723--1738}
}

@article{mackinnon1985some,
  title     = {Some heteroskedasticity-consistent covariance matrix estimators with improved finite sample properties},
  author    = {MacKinnon, James G and White, Halbert},
  journal   = {Journal of econometrics},
  volume    = {29},
  number    = {3},
  pages     = {305--325},
  year      = {1985},
  publisher = {Elsevier}
}

@article{fan1998efficient,
  title={Efficient estimation of conditional variance functions in stochastic regression},
  author={Fan, Jianqing and Yao, Qiwei},
  journal={Biometrika},
  volume={85},
  number={3},
  pages={645--660},
  year={1998},
  publisher={Oxford University Press}
}

@article{hall1989variance,
  title={Variance function estimation in regression: the effect of estimating the mean},
  author={Hall, Peter and Carroll, Raymond J},
  journal={Journal of the Royal Statistical Society Series B: Statistical Methodology},
  volume={51},
  number={1},
  pages={3--14},
  year={1989},
  publisher={Oxford University Press}
}

@article{muller1987estimation,
  title={Estimation of heteroscedasticity in regression analysis},
  author={M{\"u}ller, Hans-Georg and Stadtm{\"u}ller, Ulrich},
  journal={The Annals of Statistics},
  pages={610--625},
  year={1987},
  publisher={JSTOR}
}

@article{mcinnes2018umap,
  title={Umap: Uniform manifold approximation and projection for dimension reduction},
  author={McInnes, Leland and Healy, John and Melville, James},
  journal={arXiv preprint arXiv:1802.03426},
  year={2018}
}

@inproceedings{
dosovitskiy2021an,
title={An Image is Worth 16x16 Words: Transformers for Image Recognition at Scale},
author={Alexey Dosovitskiy and Lucas Beyer and Alexander Kolesnikov and Dirk Weissenborn and Xiaohua Zhai and Thomas Unterthiner and Mostafa Dehghani and Matthias Minderer and Georg Heigold and Sylvain Gelly and Jakob Uszkoreit and Neil Houlsby},
booktitle={International Conference on Learning Representations},
year={2021},
url={https://openreview.net/forum?id=YicbFdNTTy}
}

@inproceedings{maas2011learning,
  title={Learning word vectors for sentiment analysis},
  author={Maas, Andrew and Daly, Raymond E and Pham, Peter T and Huang, Dan and Ng, Andrew Y and Potts, Christopher},
  booktitle={Proceedings of the 49th annual meeting of the association for computational linguistics: Human language technologies},
  pages={142--150},
  year={2011}
}

@article{robbins1951stochastic,
  title     = {A stochastic approximation method},
  author    = {Robbins, Herbert and Monro, Sutton},
  journal   = {The annals of mathematical statistics},
  pages     = {400--407},
  year      = {1951},
  publisher = {JSTOR}
}

@article{sherman1950adjustment,
  title     = {Adjustment of an inverse matrix corresponding to a change in one element of a given matrix},
  author    = {Sherman, Jack and Morrison, Winifred J},
  journal   = {The Annals of Mathematical Statistics},
  volume    = {21},
  number    = {1},
  pages     = {124--127},
  year      = {1950},
  publisher = {JSTOR}
}

@inproceedings{
perezlebel2023beyond,
title={Beyond calibration: estimating the grouping loss of modern neural networks},
author={Alexandre Perez-Lebel and Marine Le Morvan and Gael Varoquaux},
booktitle={The Eleventh International Conference on Learning Representations },
year={2023},
url={https://openreview.net/forum?id=6w1k-IixnL8}
}

@article{breiman1996bagging,
  title     = {Bagging predictors},
  author    = {Breiman, Leo},
  journal   = {Machine learning},
  volume    = {24},
  number    = {2},
  pages     = {123--140},
  year      = {1996},
  publisher = {Springer}
}

@article{rivals2000construction,
  title     = {Construction of confidence intervals for neural networks based on least squares estimation},
  author    = {Rivals, Isabelle and Personnaz, L{\'e}on},
  journal   = {Neural Networks},
  volume    = {13},
  number    = {4-5},
  pages     = {463--484},
  year      = {2000},
  publisher = {Elsevier}
}

@inproceedings{grosse2016kronecker,
  title        = {A kronecker-factored approximate fisher matrix for convolution layers},
  author       = {Grosse, Roger and Martens, James},
  booktitle    = {International Conference on Machine Learning},
  pages        = {573--582},
  year         = {2016},
  organization = {PMLR}
}

@article{heskes2000natural,
  title     = {On “natural” learning and pruning in multilayered perceptrons},
  author    = {Heskes, Tom},
  journal   = {Neural Computation},
  volume    = {12},
  number    = {4},
  pages     = {881--901},
  year      = {2000},
  publisher = {MIT Press}
}

@article{suykens1999least,
  title     = {Least squares support vector machine classifiers},
  author    = {Suykens, Johan AK and Vandewalle, Joos},
  journal   = {Neural processing letters},
  volume    = {9},
  number    = {3},
  pages     = {293--300},
  year      = {1999},
  publisher = {Springer}
}

@article{fox2006effect,
  title     = {Effect displays for multinomial and proportional-odds logit models},
  author    = {Fox, John and Andersen, Robert},
  journal   = {Sociological Methodology},
  volume    = {36},
  number    = {1},
  pages     = {225--255},
  year      = {2006},
  publisher = {Wiley Online Library}
}

@article{metropolis1949monte,
  title     = {The monte carlo method},
  author    = {Metropolis, Nicholas and Ulam, Stanislaw},
  journal   = {Journal of the American statistical association},
  volume    = {44},
  number    = {247},
  pages     = {335--341},
  year      = {1949},
  publisher = {Taylor \& Francis}
}

@article{kingma2014adam,
  title   = {Adam: A method for stochastic optimization},
  author  = {Kingma, Diederik P},
  journal = {arXiv preprint arXiv:1412.6980},
  year    = {2014}
}

@article{pregibon1981logistic,
  title     = {Logistic regression diagnostics},
  author    = {Pregibon, Daryl},
  journal   = {The annals of statistics},
  volume    = {9},
  number    = {4},
  pages     = {705--724},
  year      = {1981},
  publisher = {Institute of Mathematical Statistics}
}

@inproceedings{koh2017understanding,
  title        = {Understanding black-box predictions via influence functions},
  author       = {Koh, Pang Wei and Liang, Percy},
  booktitle    = {International conference on machine learning},
  pages        = {1885--1894},
  year         = {2017},
  organization = {PMLR}
}

@article{lakshminarayanan2017simple,
  title={Simple and scalable predictive uncertainty estimation using deep ensembles},
  author={Lakshminarayanan, Balaji and Pritzel, Alexander and Blundell, Charles},
  journal={Advances in neural information processing systems},
  volume={30},
  year={2017}
}

@article{hullermeier2021aleatoric,
  title     = {Aleatoric and epistemic uncertainty in machine learning: An introduction to concepts and methods},
  author    = {H{\"u}llermeier, Eyke and Waegeman, Willem},
  journal   = {Machine learning},
  volume    = {110},
  number    = {3},
  pages     = {457--506},
  year      = {2021},
  publisher = {Springer}
}

@inproceedings{wang2025better,
  title     = {Better Training Data Attribution via Better Inverse Hessian-Vector Products},
  author    = {Andrew Wang and Elisa Nguyen and Runshi Yang and Juhan Bae and Sheila A. McIlraith and Roger Baker Grosse},
  booktitle = {The Thirty-ninth Annual Conference on Neural Information Processing Systems},
  year      = {2025}
}

@article{tibshirani_comparison_1996,
  title    = {A comparison of some error estimates for neural network models},
  volume   = {8},
  issn     = {0899-7667},
  url      = {https://doi.org/10.1162/neco.1996.8.1.152},
  doi      = {10.1162/neco.1996.8.1.152},
  number   = {1},
  urldate  = {2026-01-22},
  journal  = {Neural Comput.},
  author   = {Tibshirani, Robert},
  month    = jan,
  year     = {1996},
  pages    = {152--163}
}

@article{amari_natural_1998,
  title    = {Natural {Gradient} {Works} {Efficiently} in {Learning}},
  volume   = {10},
  issn     = {0899-7667},
  url      = {https://ieeexplore.ieee.org/abstract/document/6790500},
  doi      = {10.1162/089976698300017746},
  number   = {2},
  urldate  = {2025-10-06},
  journal  = {Neural Computation},
  author   = {Amari, Shun-ichi},
  month    = feb,
  year     = {1998},
  pages    = {251--276}
}

@inproceedings{liu_rotate_2018,
  title     = {Rotate your {Networks}: {Better} {Weight} {Consolidation} and {Less} {Catastrophic} {Forgetting}},
  doi       = {10.1109/ICPR.2018.8545895},
  booktitle = {2018 24th {International} {Conference} on {Pattern} {Recognition} ({ICPR})},
  author    = {Liu, Xialei and Masana, Marc and Herranz, Luis and Van de Weijer, Joost and López, Antonio M. and Bagdanov, Andrew D.},
  year      = {2018},
  pages     = {2262--2268}
}

@article{schmitt_general_2025,
  title     = {General {Uncertainty} {Estimation} with {Delta} {Variances}},
  volume    = {39},
  copyright = {Copyright (c) 2025 Association for the Advancement of Artificial Intelligence},
  issn      = {2374-3468},
  url       = {https://ojs.aaai.org/index.php/AAAI/article/view/34238},
  doi       = {10.1609/aaai.v39i19.34238},
  language  = {en},
  number    = {19},
  urldate   = {2026-01-29},
  journal   = {Proceedings of the AAAI Conference on Artificial Intelligence},
  author    = {Schmitt, Simon and Shawe-Taylor, John and Hasselt, Hado van},
  month     = apr,
  year      = {2025},
  pages     = {20318--20328}
}

@article{grosse2021taylor,
  title  = {Taylor approximations},
  author = {Grosse, Roger},
year = {2021}
}

@article{nilsen_epistemic_2022,
  title    = {Epistemic uncertainty quantification in deep learning classification by the {Delta} method},
  volume   = {145},
  issn     = {0893-6080},
  url      = {https://www.sciencedirect.com/science/article/pii/S0893608021004056},
  doi      = {https://doi.org/10.1016/j.neunet.2021.10.014},
  journal  = {Neural Networks},
  author   = {Nilsen, Geir K. and Munthe-Kaas, Antonella Z. and Skaug, Hans J. and Brun, Morten},
  year     = {2022},
  pages    = {164--176}
}

@inproceedings{gal_dropout_2016,
  title      = {Dropout as a {Bayesian} {Approximation}: {Representing} {Model} {Uncertainty} in {Deep} {Learning}},
  issn       = {1938-7228},
  shorttitle = {Dropout as a {Bayesian} {Approximation}},
  url        = {https://proceedings.mlr.press/v48/gal16.html},
  language   = {en},
  urldate    = {2026-01-29},
  booktitle  = {Proceedings of {The} 33rd {International} {Conference} on {Machine} {Learning}},
  publisher  = {PMLR},
  author     = {Gal, Yarin and Ghahramani, Zoubin},
  month      = jun,
  year       = {2016},
  pages      = {1050--1059}
}

@inproceedings{ritter_scalable_2018,
    title = {A {Scalable} {Laplace} {Approximation} for {Neural} {Networks}},
    url = {https://openreview.net/forum?id=Skdvd2xAZ},
    language = {en},
    urldate = {2026-07-28},
    author = {Ritter, Hippolyt and Botev, Aleksandar and Barber, David},
    month = feb,
    year = {2018},
}

@inproceedings{botev_practical_2017,
    title = {Practical {Gauss}-{Newton} {Optimisation} for {Deep} {Learning}},
    issn = {2640-3498},
    url = {https://proceedings.mlr.press/v70/botev17a.html},
    language = {en},
    urldate = {2026-07-28},
    booktitle = {Proceedings of the 34th {International} {Conference} on {Machine} {Learning}},
    publisher = {PMLR},
    author = {Botev, Aleksandar and Ritter, Hippolyt and Barber, David},
    month = jul,
    year = {2017},
    pages = {557--565},
}

@misc{immer_improving_2021,
    title = {Improving predictions of {Bayesian} neural nets via local linearization},
    url = {http://arxiv.org/abs/2008.08400},
    doi = {10.48550/arXiv.2008.08400},
    urldate = {2026-07-16},
    publisher = {arXiv},
    author = {Immer, Alexander and Korzepa, Maciej and Bauer, Matthias},
    month = feb,
    year = {2021},
    note = {arXiv:2008.08400 [stat.ML]},
}

@inproceedings{daxberger_laplace_2021,
    title = {Laplace {Redux} - {Effortless} {Bayesian} {Deep} {Learning}},
    url = {https://openreview.net/forum?id=gDcaUj4Myhn},
    language = {en},
    urldate = {2026-07-28},
    author = {Daxberger, Erik and Kristiadi, Agustinus and Immer, Alexander and Eschenhagen, Runa and Bauer, Matthias and Hennig, Philipp},
    month = nov,
    year = {2021},
}

\newpage
\appendix
\onecolumn
\section{Proofs}
\label{app:proofs}

\subsection{Fixed design variance with homoscedastic noise}

\begin{align}
\var_{\boldsymbol{\varepsilon}}\left(\hat{y}\left(\bm{x}_{\text{te}}\right)|X\right)&=\var_{\boldsymbol{\varepsilon}}\left(\bm{x}_{\text{te}}^{\top}\hat{\Sigma}_{\lambda}^{-1}X^{\top}\boldsymbol{\varepsilon}|X\right)\\
&=\bm{x}_{\text{te}}^{\top}\hat{\Sigma}_{\lambda}^{-1}X^{\top}\var_{\boldsymbol{\varepsilon}}\left(\boldsymbol{\varepsilon}\right)X\hat{\Sigma}_{\lambda}^{-1}\bm{x}_{\text{te}}  & \text{as $X$ and $\bm{x}_{\text{te}}$ are independent of $\boldsymbol{\varepsilon}$} \\
&=\sigma^{2}\bm{x}_{\text{te}}^{\top}\hat{\Sigma}_{\lambda}^{-1}\hat{\Sigma}\hat{\Sigma}_{\lambda}^{-1}\bm{x}_{\text{te}} & \text{as $\boldsymbol{\varepsilon}\sim\mathcal{N}(0,\sigma^2)$}
\label{eq:proof-fixed-design-variance}
\end{align}

\subsection{Non-parametric estimator of the variance}
The JackKnife (JK) estimator of the variance
of $\hat{y}\left(\bm{x}_{\text{te}}\right)$ uses leave-one-out estimates of the target quantity $\hat{y}\left(\bm{x}_{\text{te}}\right)^{(-i)}$, which denotes the test prediction, had we trained the model on all training examples except the $i^\text{th}$ one. It builds upon the intuition that the distribution of samples $\hat{y}\left(\bm{x}_{\text{te}}\right)^{(-i)}$ for different $i$s resembles the distribution of $\hat{y}\left(\bm{x}_{\text{te}},X,\boldsymbol{\varepsilon}\right)$ for different samples $\left(X,\boldsymbol{\varepsilon}\right)$ (i.e. different training sets). It is non-parametric as it does not require one to specify a particular parametric form for the distribution.

The estimator of the variance of $\hat{y}\left(\bm{x}_{\text{te}}\right)$ is given by:

\begin{equation}\hat{\text{var}}_{\text{JK}}\left(\hat{y}\left(\bm{x}_{\text{te}}\right)\right)=\frac{1}{n(n-1)}\sum_{i=1}^{n}\left(\bar{y}\left(\bm{x}_{\text{te}}\right)-\bar{y}\left(\bm{x}_{\text{te}}\right)^{(-i)}\right)^{2}\label{eq:app_jk_general}\end{equation} 

where $\bar{y}\left(\bm{x}_{\text{te}}\right)^{(-i)}=n\hat{y}\left(\bm{x}_{\text{te}}\right)-(n-1)\hat{y}\left(\bm{x}_{\text{te}}\right)^{(-i)}$ are called the jackknife pseudo-values and $\bar{y}\left(\bm{x}_{\text{te}}\right)=\frac{1}{n}\sum_{i=1}^{n}\bar{y}\left(\bm{x}_{\text{te}}\right)^{(-i)}$ is their average \citep{tukey1958bias}.

\subsubsection{Case 1: linear regression}

In linear regression, there exists a simple closed-form formula for $\hat{\bm{w}}^{(-i)}$ using the Sherman-Morrison formula \citep{sherman1950adjustment}:

\begin{equation*}\hat{\bm{w}}^{(-i)}=\hat{\bm{w}}-\frac{1}{1-h_{ii}}\hat{\Sigma}_{\lambda}^{-1}\bm{x}_{i}\hat{e}_i\end{equation*}

where $\hat{e}_i=y_i-\hat{y}(\bm{x}_i)$ is the residual at the training point $(\bm{x}_i,y_i)$ and $h_{ii}=\bm{x}_i^{\top}\hat{\Sigma}_{\lambda}^{-1}\bm{x}_i$ is the leverage of the training point $\bm{x}_i$, we note $\hat{u}_i=\frac{\hat{e}_i}{1-h_{ii}}$.

This leads to the following leave-one-out change of the test prediction:

\begin{equation}
\hat{y}\left(\bm{x}_{\text{te}}\right)-\hat{y}^{\left(-i\right)}\left(\bm{x}_{\text{te}}\right)=\frac{1}{1-h_{ii}}\bm{x}_{\text{te}}^{\top}\hat{\Sigma}_{\lambda}^{-1}\bm{x}_{i}\hat{e}_{i}=\bm{x}_{\text{te}}^{\top}\hat{\Sigma}_{\lambda}^{-1}\bm{x}_{i}\hat{u}_{i}
\label{eq:loo_lin}\end{equation}

In order to compute Eq. \ref{eq:app_jk_general}, we first simplify the pseudo-values average:
\begin{align}
    \bar{y}\left(\bm{x}_{\text{te}}\right)&=\frac{1}{n}\sum_{i=1}^{n}\bar{y}\left(\bm{x}_{\text{te}}\right)^{(-i)}\\&=\frac{1}{n}\sum_{i=1}^{n}\left(n\hat{y}\left(\bm{x}_{\text{te}}\right)-(n-1)\hat{y}\left(\bm{x}_{\text{te}}\right)^{(-i)}\right)\\&=\frac{1}{n}\sum_{i=1}^{n}\left(\hat{y}\left(\bm{x}_{\text{te}}\right)+(n-1)\hat{y}\left(\bm{x}_{\text{te}}\right)-(n-1)\hat{y}\left(\bm{x}_{\text{te}}\right)^{(-i)}\right)\\&=\hat{y}\left(\bm{x}_{\text{te}}\right)-\frac{n-1}{n}\sum_{i=1}^{n}\left(\hat{y}\left(\bm{x}_{\text{te}}\right)-\hat{y}\left(\bm{x}_{\text{te}}\right)^{(-i)}\right)\\\intertext{using Eq. \ref{eq:loo_lin}:}%
&=\hat{y}\left(\bm{x}_{\text{te}}\right)-\frac{n-1}{n}\bm{x}_{\text{te}}^\top\hat{\Sigma}_{\lambda}^{-1}\sum_{i=1}^{n}\bm{x}_{i}\hat{u}_{i}\\&=\hat{y}\left(\bm{x}_{\text{te}}\right)-\frac{n-1}{n}\bm{x}_{\text{te}}^\top\hat{\Sigma}_{\lambda}^{-1}X^{\top}\hat{\bm{u}}
\end{align}

Similarly for the individual pseudo-values:

\begin{align}
\bar{y}\left(\bm{x}_{\text{te}}\right)^{(-i)}&=n\hat{y}\left(\bm{x}_{\text{te}}\right)-(n-1)\hat{y}\left(\bm{x}_{\text{te}}\right)^{(-i)}\\&=\hat{y}\left(\bm{x}_{\text{te}}\right)+(n-1)\hat{y}\left(\bm{x}_{\text{te}}\right)-(n-1)\hat{y}\left(\bm{x}_{\text{te}}\right)^{(-i)}\\&=\hat{y}\left(\bm{x}_{\text{te}}\right)+(n-1)\left(\hat{y}\left(\bm{x}_{\text{te}}\right)-\hat{y}\left(\bm{x}_{\text{te}}\right)^{(-i)}\right)\\\intertext{using Eq. \ref{eq:loo_lin}:}&=\hat{y}\left(\bm{x}_{\text{te}}\right)+(n-1)\bm{x}_{\text{te}}^\top\hat{\Sigma}_{\lambda}^{-1}\bm{x}_{i}\hat{u}_{i}
\end{align}

Finally, the difference between the pseudo-values and their average is:

\begin{align}\bar{y}\left(\bm{x}_{\text{te}}\right)-\bar{y}\left(\bm{x}_{\text{te}}\right)^{(-i)}&=(n-1)\bm{x}_{\text{te}}^{\top}\hat{\Sigma}_{\lambda}^{-1}\left(\bm{x}_{i}\hat{u}_{i}-\frac{1}{n}\sum_{i=1}^{n}\bm{x}_{i}\hat{u}_{i}\right)\\&=(n-1)\bm{x}_{\text{te}}^{\top}\hat{\Sigma}_{\lambda}^{-1}\left(\bm{x}_{i}\hat{u}_{i}-\frac{1}{n}X^{\top}\hat{\bm{u}}\right)
\end{align}

Thus, getting back to Eq. \ref{eq:app_jk_general}, the jackknife estimator of the variance of $\hat{y}\left(\bm{x}_{\text{te}}\right)$ gives:

\begin{align}\hat{\text{var}}_{\text{JK}}\left(\hat{y}\left(\bm{x}_{\text{te}}\right)\right)&=\frac{(n-1)^{2}}{n(n-1)}\bm{x}_{\text{te}}^{\top}\hat{\Sigma}_{\lambda}^{-1}\left(\sum_{i=1}^{n}\left(\bm{x}_{i}\hat{u}_{i}-\frac{1}{n}X^{\top}\hat{\bm{u}}\right)\left(\bm{x}_{i}\hat{u}_{i}-\frac{1}{n}X^{\top}\hat{\bm{u}}\right)^{\top}\right)\hat{\Sigma}_{\lambda}^{-1}\bm{x}_{\text{te}}\\&=\frac{n-1}{n}\bm{x}_{\text{te}}^{\top}\hat{\Sigma}_{\lambda}^{-1}\left(\sum_{i=1}^{n}\bm{x}_{i}\hat{u}_{i}^{2}\bm{x}_{i}^{\top}-\frac{1}{n^{2}}\sum_{i=1}^{n}X^{\top}\hat{\bm{u}}\hat{\bm{u}}^{\top}X\right)\hat{\Sigma}_{\lambda}^{-1}\bm{x}_{\text{te}}\\&=\frac{n-1}{n}\bm{x}_{\text{te}}^{\top}\hat{\Sigma}_{\lambda}^{-1}\left(X^{\top}\text{diag}\left(\hat{\bm{u}}^{2}\right)X-\frac{1}{n}X^{\top'}\hat{\bm{u}}\hat{\bm{u}}^{\top}X\right)\hat{\Sigma}_{\lambda}^{-1}\bm{x}_{\text{te}}\\&=\frac{n-1}{n}\bm{x}_{\text{te}}^{\top}\hat{\Sigma}_{\lambda}^{-1}X^{\top}\left(\text{diag}\left(\hat{\bm{u}}^{2}\right)-\frac{1}{n}\hat{\bm{u}}\hat{\bm{u}}^{\top}\right)X\hat{\Sigma}_{\lambda}^{-1}\bm{x}_{\text{te}}
\end{align}

This closely resembles the estimator of the mean variance of the risk of \citet{mackinnon1985some}, but here applied to a single test prediction $\bm{x}_{\text{te}}$.

We additionnally drop the term $\frac{1}{n}\hat{\bm{u}}\hat{\bm{u}}^{\top}$ which shrinks as $\frac{1}{n}$. This estimator, named HC3, is equivalent to assuming that there is no bias in the jackknife estimator $\bar{\bm{w}}=\hat{\bm{w}}$ \citep{mackinnon1985some}.

Finally, dropping the constant $\frac{n-1}{n}$, the jackknife estimator of the variance at a test point $\bm{x}_{\text{te}}$ is:

\begin{equation}
\hat{\var}_{\text{JK}}\left(\hat{y}(\bm{x}_{\text{te}})\right)=\bm{x}_{\text{te}}^{\top}\hat{\Sigma}_{\lambda}^{-1}X^{\top}\diag(\hat{\bm{u}}^2)X\hat{\Sigma}_{\lambda}^{-1}\bm{x}_{\text{te}}
\end{equation}

As an aside, we can equivalently use the infinitesimal jackknife \cite{jaeckel1972infinitesimal} which recovers the usual SandWich (SW) estimator \citep{white1980heteroskedasticity} or HC estimator \citep{mackinnon1985some}:

\begin{equation}
\hat{\var}_{\text{SW}}\left(\hat{y}\left(\bm{x}_{\text{te}}\right)\right)=\bm{x}_{\text{te}}^{\top}\hat{\Sigma}_{\lambda}^{-1}X^{\top}\diag(\hat{\bm{e}}^2)X\hat{\Sigma}_{\lambda}^{-1}\bm{x}_{\text{te}}
\end{equation}

\subsubsection{Case 2: non-linear regression (deep learning)}

This is similar to the case of linear regression, by replacing $X$ by $\Phi_{\bm{w}}$, $\hat{\Sigma}_\lambda$ by $F_{\bm{\hat{w}}\lambda}$, and $\boldsymbol{\varepsilon}$ by $\bm{r}$.

\subsection{Linearization}

\begin{align}\left\Vert \bm{y}-f_{\bm{w}}\left(X\right)\right\Vert ^{2}+\lambda\left\Vert \bm{w}\right\Vert ^{2}&=\left\Vert \underbrace{\bm{y}-f_{\hat{\bm{w}}}\left(X\right)}_{:=\text{ pseudo-responses }\bm{r}}-\Phi_{\bm{\hat{w}}}\bm{\Delta}\bm{w}+o\left(\left\Vert \bm{\Delta}\bm{w}\right\Vert \right)\right\Vert ^{2}+\lambda\left\Vert \bm{\hat{w}}+\bm{\Delta w} \right\Vert ^{2}\\&=\left\Vert \bm{r}-\Phi_{\bm{\hat{w}}}\bm{\Delta}\bm{w}+o\left(\left\Vert \bm{\Delta}\bm{w}\right\Vert \right)\right\Vert ^{2}+\lambda\left\Vert \bm{\hat{w}}+\bm{\Delta w} \right\Vert ^{2}\\&=\left\Vert \bm{r}-\Phi_{\bm{\hat{w}}}\bm{\Delta}\bm{w}\right\Vert ^{2}+o\left(\left\Vert \bm{\Delta}\bm{w}\right\Vert \right)+\lambda\left\Vert \bm{\Delta w} \right\Vert ^{2}+\lambda\left\Vert \bm{\hat{w}} \right\Vert ^{2}+2\lambda\bm{\hat{w}}^{\top}\bm{\Delta w}\\&=\bm{\Delta}\bm{w}^{\top}\left(\Phi_{\bm{\hat{w}}}^{\top}\Phi_{\bm{\hat{w}}}+\lambda I\right)\bm{\Delta}\bm{w}+2\left(\lambda\bm{\hat{w}}-\bm{r}^{\top}\Phi_{\bm{\hat{w}}}\right)\bm{\Delta}\bm{w}+o\left(\left\Vert \bm{\Delta}\bm{w}\right\Vert \right)+C
\end{align}

takes it minimum (up to $o\left(\left\Vert \bm{\Delta}\bm{w}\right\Vert \right))$ for

\begin{align}
\bm{\Delta}\bm{w}&=\left(\Phi_{\bm{\hat{w}}}^{\top}\Phi_{\bm{\hat{w}}}+\lambda I\right)^{-1}\left(\Phi_{\bm{\hat{w}}}^{\top}\bm{r}-\lambda\bm{\hat{w}}\right)
\end{align}

\subsection{EKFAC}
\label{app:ekfac}

\paragraph{\texttt{HeC3} estimator with EKFAC:}

\begin{align}
\hat{\var}_{\text{JK}}\left(f_{\hat{\bm{w}}}(\bm{x}_{\text{te}})\right)&=\bm{\phi}(\bm{x}_{\text{te}})^{\top}\left(\Phi^{\top}_{\hat{\bm{w}}}\Phi_{\hat{\bm{w}}}+\lambda I\right)^{-1}\left(\Phi^{\top}_{\hat{\bm{w}}}\diag(\hat{\bm{u}}^2)\Phi_{\hat{\bm{w}}}\right)\left(\Phi^{\top}_{\hat{\bm{w}}}\Phi_{\hat{\bm{w}}}+\lambda I\right)^{-1}\bm{\phi}(\bm{x}_{\text{te}})\\
\intertext{We estimate $\Phi^{\top}_{\hat{\bm{w}}}\Phi_{\hat{\bm{w}}}$ with EKFAC: let $U$ be an orthogonal matrix such that $U^{\top}=U^{-1}$, and $S^{*}$ a diagonal matrix such that $S^{*}=\diag(\bm{s}^{*})$, we have: $\left(\Phi^{\top}_{\hat{\bm{w}}}\Phi_{\hat{\bm{w}}}+\lambda I\right)=U(S^{*} + \lambda I)U^{\top}$ so:}
&=\bm{\phi}(\bm{x}_{\text{te}})^{\top}\left(U(S^{*} + \lambda I)U^{\top} \right)^{-1}\left(\Phi^{\top}_{\hat{\bm{w}}}\diag(\hat{\bm{u}}^2)\Phi_{\hat{\bm{w}}}\right)\left(U(S^{*} + \lambda I)U^{\top}\right)^{-1}\bm{\phi}(\bm{x}_{\text{te}})\\
\intertext{Yet $\Phi^{\top}_{\hat{\bm{w}}}\diag(\hat{\bm{u}}^2)\Phi_{\hat{\bm{w}}}$ is not simultaneously diagonalisable with $\Phi^{\top}_{\hat{\bm{w}}}\Phi_{\hat{\bm{w}}}$, meaning diagonalisable in the same eigen basis $U$, as both matrices do not commute. We approximate $\Phi^{\top}_{\hat{\bm{w}}}\diag(\hat{\bm{u}}^2)\Phi_{\hat{\bm{w}}}$ with $UD^{*}U^T$ with the best diagonal matrix $D^{*}=\diag\left(\bm{d}^{*}\right)$ that minimized the frobenius norm of the difference of the approximated matrix and the original matrix \citep{george2018fast}:}
&=\bm{\phi}(\bm{x}_{\text{te}})^{\top}\left(U(S^{*} + \lambda I)U^{\top} \right)^{-1}\left(UD^{*}U^{\top}\right)\left(U(S^{*} + \lambda I)U^{\top}\right)^{-1}\bm{\phi}(\bm{x}_{\text{te}})\\
&=\bm{\phi}(\bm{x}_{\text{te}})^{\top}\left(U^{-\top}(S^{*} + \lambda I)^{-1}U^{-1} \right)\left(UD^{*}U^{\top}\right)\left(U^{-\top}(S^{*} + \lambda I)^{-1}U^{-1}\right)\bm{\phi}(\bm{x}_{\text{te}})\\
\intertext{As $U^{-\top}=\left(U^{\top}\right)^{-1}=U$:}
&=\bm{\phi}(\bm{x}_{\text{te}})^{\top}\left(U(S^{*} + \lambda I)^{-1}U^{-1} UD^{*}U^{\top}U^{-\top}(S^{*} + \lambda I)^{-1}U^{\top}\right)\bm{\phi}(\bm{x}_{\text{te}})\\
&=\bm{\phi}(\bm{x}_{\text{te}})^{\top}\left(U(S^{*} + \lambda I)^{-1}D^{*}(S^{*} + \lambda I)^{-1}U^{\top}\right)\bm{\phi}(\bm{x}_{\text{te}})\\
&=\bm{\phi}(\bm{x}_{\text{te}})^{\top}\left(U\diag\left(\frac{\bm{d}^{*}}{(\bm{s}^{*}+\lambda)^2}\right)U^{\top}\right)\bm{\phi}(\bm{x}_{\text{te}})\\
&=\left\Vert\left(\frac{\bm{d}^*}{(\bm{s}^*+\lambda)^2}\right)^{\frac{1}{2}}U^{\top}\bm{\phi}(\bm{x}_{\text{te}})\right\Vert_2^2\\
\intertext{If we don't assume that both covariances are simultaneously diagonalizable, we can still leverage that for EKFAC, $U=U_A\otimes U_B$. We note as $VD^{*}V^{\top}$ the EKFAC approximation of $\Phi^{\top}_{\hat{\bm{w}}}\diag(\hat{\bm{u}}^2)\Phi_{\hat{\bm{w}}}$ with $V=V_A\otimes V_B$:}
&=\bm{\phi}(\bm{x}_{\text{te}})^{\top}\left(U(S^{*} + \lambda I)U^{\top} \right)^{-1}\left(VD^{*}V^{\top}\right)\left(U(S^{*} + \lambda I)U^{\top}\right)^{-1}\bm{\phi}(\bm{x}_{\text{te}})\\
&=\bm{\phi}(\bm{x}_{\text{te}})^{\top}\left(U^{-\top}(S^{*} + \lambda I)^{-1}U^{-1} \right)\left(VD^{*}V^{\top}\right)\left(U^{-\top}(S^{*} + \lambda I)^{-1}U^{-1}\right)\bm{\phi}(\bm{x}_{\text{te}})\\
&=\bm{\phi}(\bm{x}_{\text{te}})^{\top}\left(U(S^{*} + \lambda I)^{-1}U^{-1} VD^{*}V^{\top}U^{-\top}(S^{*} + \lambda I)^{-1}U^{\top}\right)\bm{\phi}(\bm{x}_{\text{te}})\\
\intertext{As $(A\otimes B)(C \otimes D)=(AB)\otimes (CD)$ and $(A\otimes B)^{\top}=(A^{\top}\otimes B^{\top})$ }
&=\bm{\phi}(\bm{x}_{\text{te}})^{\top}\left(U(S^{*} + \lambda I)^{-1}\left((U_A^{\top}V_A)\otimes(U_B^{\top}V_B)\right)D^{*}\left((V_A^{\top}U_A)\otimes(V_B^{\top}U_B)\right)(S^{*} + \lambda I)^{-1}U^{\top}\right)\bm{\phi}(\bm{x}_{\text{te}})\\
&=\left\Vert {D^*}^{\frac{1}{2}}\left((V_A^{\top}U_A)\otimes(V_B^{\top}U_B)\right)(S^* + \lambda I)^{-1}U^{\top}\bm{\phi}(\bm{x}_{\text{te}})\right\Vert_2^2
\end{align}

Note that we can do similar derivations for the infinitesimal jackknife, using $\hat{\bm{e}}$ instead of $\hat{\bm{u}}$.

\paragraph{MLE estimator with EKFAC:}

\begin{align}
\hat{\var}_{\text{MLE}}\left(f_{\hat{\bm{w}}}(\bm{x}_{\text{te}})\right)&=\bm{\phi}(\bm{x}_{\text{te}})^{\top}\left(\Phi^{\top}_{\hat{\bm{w}}}\Phi_{\hat{\bm{w}}}+\lambda I\right)^{-1}\Phi^{\top}_{\hat{\bm{w}}}\Phi_{\hat{\bm{w}}}\left(\Phi^{\top}_{\hat{\bm{w}}}\Phi_{\hat{\bm{w}}}+\lambda I\right)^{-1}\bm{\phi}(\bm{x}_{\text{te}})\\
&=\bm{\phi}(\bm{x}_{\text{te}})^{\top}U(S^{*}+\lambda I)^{-1}S^{*}(S^{*}+\lambda I)^{-1}U^{\top}\bm{\phi}(\bm{x}_{\text{te}})\\
&=\bm{\phi}(\bm{x}_{\text{te}})^{\top}U\diag\left(\frac{\bm{s}^{*}}{(\bm{s}^{*}+\lambda)^2}\right)U^{\top}\bm{\phi}(\bm{x}_{\text{te}})\\
&=\left\Vert\left(\frac{\bm{s}^*}{(\bm{s}^*+\lambda)^2}\right)^{\frac{1}{2}}U^{\top}\bm{\phi}(\bm{x}_{\text{te}})\right\Vert_2^2
\end{align}

\subsection{\texttt{HeC3}/\texttt{Ho} as a linear smoother of Leave-One-Out Residuals}
\label{app:hec3_ho_kernel_smoothing}

To gain theoretical intuition into the behavior of the ratio $\texttt{HeC3}/\texttt{Ho}$, we show that it can be expressed as a linear smoother of leave-one-out residuals. This reformulation reveals that the ratio has the same algebraic structure as classical non-parametric estimators of conditional variance (a term used for aleatoric uncertainty in the econometrics literature \citep{muller1987estimation}), although the notion of locality is induced by the geometry of the linearized model rather than by the input space. 

For clarity, we present the derivation for the $\texttt{HeC0}/\texttt{Ho}$, which is more in line with the econometrics literature.

Let $a_i({\bm{x}_{\text{te}}}) = \phi(\bm{x}_i)^\top F_{\bm{\hat{w}}\lambda}^{-1} \phi(\bm{x}_{\text{te}})$ be the alignment between the tangent features of $x_i$ and $x_{\mathrm{te}}$ after rescaling by the inverse regularized Fisher Information Matrix. In particular, it defines a similarity in the geometry induced by the linearized model rather than in the original input space.

The homoscedastic estimator $\texttt{Ho}(\bm{x}_{\text{te}})$ assumes a global noise variance $\hat{\sigma}^2 = \frac{1}{n} \sum_{j=1}^n \hat{e}_j^2$ and expands as:
$$
\texttt{Ho}(\bm{x}_{\text{te}}) = \hat{\sigma}^2 \phi(\bm{x}_{\text{te}})^\top F_{\bm{\hat{w}}\lambda}^{-1} \Phi^\top \Phi F_{\bm{\hat{w}}\lambda}^{-1} \phi(\bm{x}_{\text{te}}) = \hat{\sigma}^2 \sum_{i=1}^n a_i(\bm{x}_{\text{te}})^2.
$$

Similarly, the uncorrected heteroscedastic estimator $\texttt{HeC0}(\bm{x}_{\text{te}})$ uses sample residuals $\hat{e}_i = y_i - \hat{f}(\bm{x}_i)$:
$$
\texttt{HeC0}(\bm{x}_{\text{te}}) = \phi(\bm{x}_{\text{te}})^\top F_{\bm{\hat{w}}\lambda}^{-1} \Phi^\top \text{diag}(\hat{e}^2) \Phi F_{\bm{\hat{w}}\lambda}^{-1} \phi(\bm{x}_{\text{te}}) = \sum_{i=1}^n a_i(\bm{x}_{\text{te}})^2 \hat{e}_i^2.
$$

By taking the ratio of $\texttt{HeC0}(\bm{x}_{\text{te}})$ to $\texttt{Ho}(\bm{x}_{\text{te}})$ scaled by the global variance $\hat{\sigma}^2$, we obtain:
$$
\hat{\sigma}^2 \frac{\texttt{HeC0}(\bm{x}_{\text{te}})}{\texttt{Ho}(\bm{x}_{\text{te}})} = \sum_{i=1}^n \omega_i(\bm{x}_{\text{te}}) \hat{e}_i^2, \quad \text{where } \omega_i(\bm{x}_{\text{te}}) = \frac{a_i(\bm{x}_{\text{te}})^2}{\sum_{j=1}^n a_j(\bm{x}_{\text{te}})^2}.
$$

Since $\omega_i(\bm{x}_{\text{te}}) \ge 0$ and $\sum_{i=1}^n \omega_i(\bm{x}_{\text{te}}) = 1$, the ratio takes the form of a linear smoother of squared residuals. The weights are induced by the squared alignment:
$$K(\bm{x}_i,\bm{x}_{\text{te}})=a_i(\bm{x}_{\text{te}})^2$$
The ratio has the same structure as a kernel smoother of squared residuals, where similarity is measured in tangent-feature space rather than input feature space. It is an estimator commonly used in econometrics for the conditional variance \citep{muller1987estimation,hall1989variance,fan1998efficient}.

It can also be seen as a influence weighted expectation, the normalized weights $\omega_i(\bm{x}_{\text{te}})$ define a local empirical probability measure $\mathbb{P}_{ \bm{x}_{\text{te}}} = \sum_{i=1}^n \omega_i(\bm{x}_{\text{te}}) \delta_{(\bm{x}_i, y_i)}$. The ratio is therefore the expected squared residual under a training distribution re-weighted by each sample's influence on $\bm{x}_{\text{te}}$:
    $$
    \hat{\sigma}^2 \frac{\texttt{HeC0}(\bm{x}_{\text{te}})}{\texttt{Ho}(\bm{x}_{\text{te}})} = \mathbb{E}_{(X, Y) \sim \mathbb{P}_{ \bm{x}_{\text{te}}}} \left( (Y - \hat{f}(X))^2 \right).
    $$

For deep neural networks, training residuals are often overly optimistic due
to overfitting. The \texttt{HeC3} estimator preserves the same weighting while replacing
the residuals by their jackknife (leave-one-out) corrections $\hat{u}_i = \frac{\hat{e}_i}{1 - h_{ii}}$, where $h_{ii} = \phi(\bm{x}_i)^\top F_{\bm{\hat{w}}\lambda}^{-1} \phi(\bm{x}_i)$ is the leverage. Therefore:
$$
\hat{\sigma}^2 \frac{\texttt{HeC3}(\bm{x}_{\text{te}})}{\texttt{Ho}(\bm{x}_{\text{te}})} = \sum_{i=1}^n \omega_i(\bm{x}_{\text{te}}) \hat{u}_i^2.
$$

Thus, $\texttt{HeC3}(\bm{x}_{\text{te}}) / \texttt{Ho}(\bm{x}_{\text{te}})$ directly measures the local, loo residual variance around $\bm{x}_{\text{te}}$ relative to the global variance $\hat{\sigma}^2$.

\section{Technical Appendices}\label{app:technical}

\subsection{Multiclass Classification as Regression}

In all classification experiments, we reformulate the problem as a regression task, akin to the training Least Squares Support Vector Machines \citep{suykens1999least}. For binary and multi-class classification, we one-hot encode the target variables $\bm{y}$ as $Y$ where:
  \begin{equation}
    Y_{ij}=
    \begin{cases}
    1, & \text{if}\ y_i =j \\
    -1, & \text{otherwise}
    \end{cases}
  \end{equation}
Then we train the neural network to minimize the mean-squared error between the outputs of the last layer and the targets $Y$.

For $k$-output linear regression: $Y=XW^*$ with $X$ of size $n\times d$ and $W^*$ of size $d\times k$, the estimated ridge weights are $\hat{W}=\left(X^\top X+\lambda I\right)^{-1}X^\top Y$ and residuals $E=Y-X\hat{W}$. To better match formulas from the linear regression with one output, we adopt a stacked notation where $\bm{y}$ is the $nk$ vector of targets with $\bm{y}=\text{vec}(Y)=(I_k \otimes X)\text{vec}(W^*)=(I_k \otimes X)\bm{w}^*$ with $\bm{w}^*$ the stacked $kd$ vector of weights. Assuming homoscedastic noise, the covariance of the residuals $\text{vec}(E)=\boldsymbol{\varepsilon}$ takes the following form $\Sigma_E\otimes I_n$, with $n\Sigma_E=E^\top E$, leading to the following maximum likelihood estimator of the variance:

\begin{align*}
    \var_{\boldsymbol{\varepsilon}}\left(\hat{\bm{y}}(\bm{x}_{\text{te}})|X\right)&=\var_{\boldsymbol{\varepsilon}}\left((I_k \otimes \bm{x}_{\text{te}}^{\top})\left((I_k \otimes X)^{\top}(I_k \otimes X)+\lambda I_{kd}\right)^{-1}(I_k \otimes X)^\top \boldsymbol{\varepsilon}|X\right)\\
    &=\var_{\boldsymbol{\varepsilon}}\left((I_k \otimes \bm{x}_{\text{te}}^{\top})\left(I_k\otimes\left(X^{\top}X+\lambda I_d\right)\right)^{-1}(I_k \otimes X)^\top \boldsymbol{\varepsilon}|X\right)\\
    &=\var_{\boldsymbol{\varepsilon}}\left(I_k \otimes \bm{x}_{\text{te}}^{\top}(X^{\top}X+\lambda I_d)^{-1}X^\top \boldsymbol{\varepsilon}|X\right)\\
    &=\left(I_k \otimes\bm{x}_{\text{te}}^{\top}(X^{\top}X+\lambda I_d)^{-1}X^\top\right)\var_{\boldsymbol{\varepsilon}}(\boldsymbol{\varepsilon})\left( I_k \otimes X(X^{\top}X+\lambda I_d)^{-1}\bm{x}_{\text{te}}  \right)\\
    &=\left(I_k \otimes\bm{x}_{\text{te}}^{\top}(X^{\top}X+\lambda I_d)^{-1}X^\top\right)\left( \Sigma_E \otimes I_n\right)\left( I_k \otimes X(X^{\top}X+\lambda I_d)^{-1}\bm{x}_{\text{te}}  \right)\\
    &=\Sigma_E \otimes \bm{x}_{\text{te}}^{\top}(X^{\top}X+\lambda I_d)^{-1}X^{\top}X(X^{\top}X+\lambda I_d)^{-1}\bm{x}_{\text{te}} 
\end{align*}

The derivation of the maximum likelihood estimator is simplified in the previous case because the inputs of the linear model $X$ are constant across classes. However, in the case of linearized networks with tangent features, the inputs correspond to the Jacobian of the output of the neural network for the corresponding class: $I_k\otimes X$ becomes $\bm{\Phi}$ the stacked $nk\times d$ tangent features, with $\bm{\phi}(\bm{x})$ is the $k\times d$ tangent features of example $\bm{x}$. Thus, the sandwich form never simplifies:

\begin{align*}
    \var_{\boldsymbol{\varepsilon}}\left(\hat{\bm{f}}(\bm{x}_{\text{te}})|X\right)
    &=\var_{\boldsymbol{\varepsilon}}\left(\bm{\phi}(\bm{x}_{\text{te}})^{\top}\left(\bm{\Phi}^{\top}\bm{\Phi}+\lambda I\right)^{-1}\bm{\Phi}^\top \boldsymbol{\varepsilon}|X\right)\\
    &=\left(\bm{\phi}(\bm{x}_{\text{te}})^{\top}\left(\bm{\Phi}^{\top}\bm{\Phi}+\lambda I\right)^{-1}\bm{\Phi}^\top \right) \var_{\boldsymbol{\varepsilon}}(\boldsymbol{\varepsilon})\left(\bm{\Phi}\left(\bm{\Phi}^{\top}\bm{\Phi}+\lambda I\right)^{-1}\bm{\phi}(\bm{x}_{\text{te}})\right)\\
    &=\left(\bm{\phi}(\bm{x}_{\text{te}})^{\top}\left(\bm{\Phi}^{\top}\bm{\Phi}+\lambda I\right)^{-1}\bm{\Phi}^\top \right)\left(\Sigma_E\otimes I_n\right)\left(\bm{\Phi}\left(\bm{\Phi}^{\top}\bm{\Phi}+\lambda I\right)^{-1}\bm{\phi}(\bm{x}_{\text{te}})\right) \\
    &=\bm{\phi}(\bm{x}_{\text{te}})^{\top}\left(\bm{\Phi}^{\top}\bm{\Phi}+\lambda I\right)^{-1}\left(\left(\left(\Sigma_E^{\frac{1}{2}}\otimes I_n\right)\bm{\Phi}\right)^\top \left(\left(\Sigma_E^{\frac{1}{2}}\otimes I_n\right)\bm{\Phi}\right)\right)\left(\bm{\Phi}^{\top}\bm{\Phi}+\lambda I\right)^{-1}\bm{\phi}(\bm{x}_{\text{te}}) \\
\end{align*}

where $\Sigma_E^{\frac{1}{2}}$ is computed from the Cholesky decomposition of $\Sigma_E$.

In a multi-class scenario, the predicted uncertainty will be a square matrix with the same size as the number of classes. To get a scalar variance, we compute the sum of the per-class variances, akin to using the trace of the covariance matrix as a generalized variance.

Even though deriving estimators to quantify the uncertainty of the predictions of a neural network after a softmax layer is possible, using, for example, the Delta method \citep{cox1995asymptotic} or the double Delta method \citep{fox2006effect}, we expect to do that in future work.

\subsection{Meta-Classes}
\label{app:meta}

For image classification datasets, constructing meaningful subpopulations of the total training population is not straightforward: unlike in tabular data, there is no obvious categorical feature we could use to achieve such a split. That's why, for MNIST and Fashion-MNIST, we use the original class labels as the categorical feature to create subpopulations. As the original class labels are no longer available, we construct a new binary classification task, by merging the original classes completely at random. Then we pick one of the original classes at random that we remove from the training set to create our missing subpopulation, and another one (different) where we assign random labels to create our noisy subpopulation.

\subsection{Hyperparameters}
\label{app:hyperparams}

For the nonlinear regression dataset, we used a one-layer MLP with 100 hidden neurons and $\tanh$ activation. It was trained with the L-BFGS \citep{nocedal1980updating,liu1989limited} optimizer at a learning rate of $0.1$ for $100$ steps.

For the \emph{spirals} dataset, we used a one-layer MLP with 20 hidden neurons and $\tanh$ activation. It was trained with the L-BFGS \citep{nocedal1980updating,liu1989limited} optimizer at a learning rate of $0.1$ for $200$ steps.

For the two tabular datasets \emph{bank-marketing} \citep{moro2014data} and \emph{california-housing} \citep{pace1997sparse}, we used a one-layer MLP with 100 hidden neurons, ReLU activations, and Layer Normalization. It was trained with the Adam optimizer \cite{kingma2014adam} for $100$ epochs on \emph{california-housing} and $50$ epochs on \emph{bank-marketing}, with a batch size of $128$.

For the \emph{MNIST} dataset \citep{LeCun2005TheMD}, we used a two-layer MLP with $1024$ hidden neurons. It was trained with the Adam optimizer \cite{kingma2014adam} with \num{1e-3} learning rate, \num{1e-2} weight decay for $20$ epochs with batch size $128$.

For the \emph{Fashion-MNIST} dataset \citep{xiao2017fashion}, we used a 4-layer CNN followed by a linear classification layer. The conv layers have channel sizes of $32$, $64$, $128$, and $256$ with $3\times3$ kernels. It was trained with the Adam optimizer \cite{kingma2014adam} with \num{1e-3} learning rate, \num{1e-2} weight decay for $20$ epochs with batch size $128$.

For the \emph{CIFAR-10} dataset \citep{krizhevsky2009learning} we used a ResNet18 \citep{he2016deep} that we trained with the Adam optimizer \cite{kingma2014adam} with \num{1e-3} learning rate, \num{1e-2} weight decay for $50$ epochs with batch size $128$. We also used a ViT \citep{dosovitskiy2021an} trained with the same optimizer and hyperparameters as the ResNet, for $100$ epochs.

For the \emph{IMDB} dataset \citep{maas2011learning} we used a BERT \citep{devlin2019bert} that we fined-tune with the Adam optimizer \cite{kingma2014adam} with \num{1e-3} learning rate, \num{1e-2} weight decay for $10$ epochs with batch size $128$.

\subsection{Comparison to other uncertainty quantification techniques}
\label{app:competitors}

We also implemented two well-known approaches to compute uncertainty for deep neural networks: Deep Ensemble \citep{lakshminarayanan2017simple}. Deep Ensemble estimates a neural network's uncertainty by training the same architecture on the same dataset with different weight initializations. As this method is computationally intensive, we limit the number of training models to $10$.

\subsection{KFE implementation}

In order to compute the EKFAC estimation of the Fisher Information matrix in an efficient manner, we need to be able to compute a Kronecker Factorization of the FIM of each layer. This factorization is known \citep{heskes2000natural} for linear layers and convolutional layers \citep{grosse2016kronecker}. For other kinds of layers, it seems difficult to imagine how a Kronecker Factorization could be done, i.e., batch norm or layer norm layers. For these layers, we resort to an exact block-diagonal approach, which we also diagonalize via its eigendecomposition. Thus, we compute neural tangent features efficiently for all layers of a ResNet, ViT, or BERT.

\subsection{Jackknife residuals and EKFAC}

The jackknife residual or Leave-One-Out residuals are:
$$\hat{u}_i=\frac{\hat{e}_i}{1-h_{ii}}$$
where $\hat{e}_i=y_i - \hat{f}(\bm{x}_i)$ is the residual and $h_{ii}=\phi(\bm{x}_i)F_{\bm{\hat{w}}\lambda}^{-1}\phi(\bm{x}_i)^{\top}$ the leverage of the training point $\bm{x}_i$.

When using the exact Fisher Information Matrix, the leverages are guaranteed to be between $0$ and $1$. Indeed, using the singular value decomposition, $\Phi=U \diag(s) V^\top$, of the tangent features, the corresponding hat matrix is:

$$H=U\diag\left(\frac{s^2}{s^2+\lambda}\right)U^\top$$

and the leverage of $\bm{x}_i$ is:

$$h_{ii}=\sum_j \bm{u}_{ij}^2\frac{s_j^2}{s_j^2 + \lambda}$$

Since $\forall i$, $0\leq\frac {s_j^2}{s_j^2 + \lambda}\leq1$ and $\sum_j \bm{u}_{ij}^2\leq1$ because the columns of U are orthonormal, the leverages are bounded between $0$ and $1$, and strictly less than $1$ for $\lambda>0$.

However, when using approximations of the FIM, such as EKFAC, there is no guarantee that these leverages are bounded between $0$ and $1$, and can make $1/(1-h_{ii})$ negative, which is why in practice, the leverages computed with EKFAC are then clipped to $0$ and $1-\epsilon$, we chose $\epsilon=\num{1e-4}$.

\section{Related works -- extended}

\paragraph{\citet{ritter_scalable_2018}} is a popular Bayesian perspective on uncertainty quantification, mixing the Laplace approximation to the posterior over weights, with the K-FAC \citep{martens2015optimizing} and KFRA \citep{botev_practical_2017} methods for efficiently approximating the Fisher Information Matrix.

\paragraph{\citet{immer_improving_2021}} show the link between the Generalized Gauss-Newton approximation to the Laplace posterior, and a linearization of the deep network predictor. They advocate for using a slightly different procedure for posterior sampling.

\paragraph{\citet{daxberger_laplace_2021}} propose a library that implements the numerous variants of the Laplace approximation and provide a comprehensive and easy to read presentation of the concepts. The Generalized Gauss-Newton/Fisher Information Matrix can be approximated using K-FAC, KFRA, or a diagonal approximation.

\paragraph{\citet{schmitt_general_2025}} aim at re-popularizing delta variance estimators in deep learning, and show that they can be derived using a Bayesian or frequentist perspective. They evaluate a delta variance estimator using a diagonal approximate of the inverse Fisher information matrix on a large scale weather forecast dataset. They also propose to learn the inverse covariance matrix coefficients, using gradient descent.

\clearpage

\section{Additional Experiments}
\label{app:xp}

\begin{figure}[H]
    \centering
    \includegraphics[width=\linewidth]{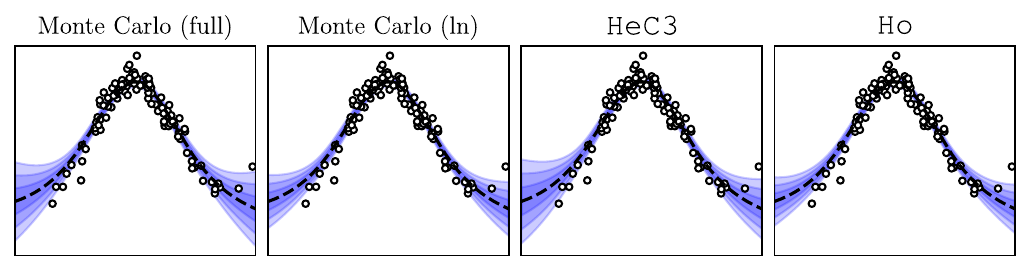}
    \includegraphics[width=\linewidth]{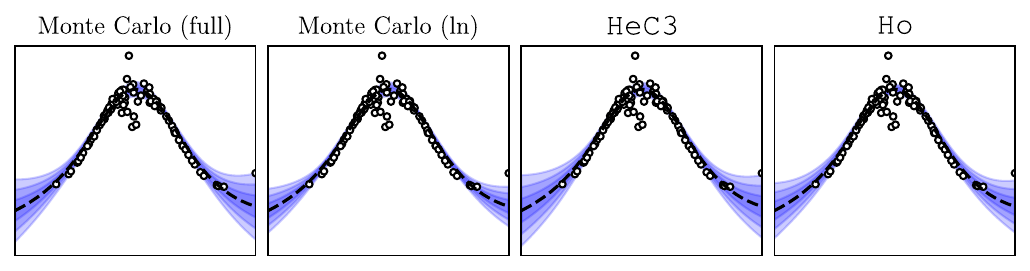}
    \includegraphics[width=\linewidth]{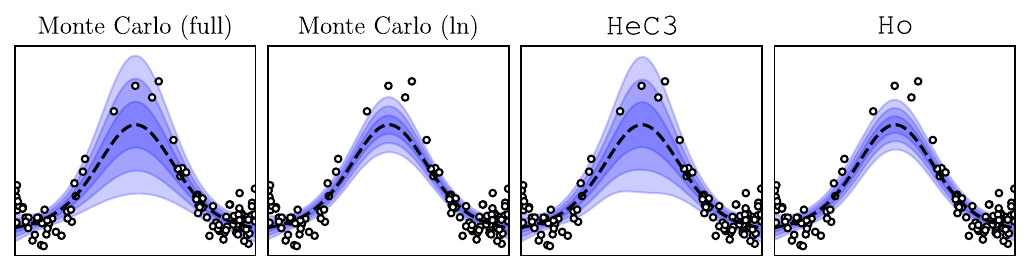}
    \caption{Nonlinear regression task with (from top to bottom) homoscedasticity, heteroscedasticity near $x\rightarrow0$, and fewer samples near $x\rightarrow0$. Estimated variance of neural networks with \texttt{HeC3} and \texttt{Ho} estimators. Ground truth variances are estimated using Monte Carlo (100 trials), by resampling a full new dataset or by adding homoscedastic label noise on a fixed dataset.}
    \label{fig:xsinx-comparison}
\end{figure}

\begin{figure}[H]
    \centering
    \includegraphics[width=0.6\linewidth]{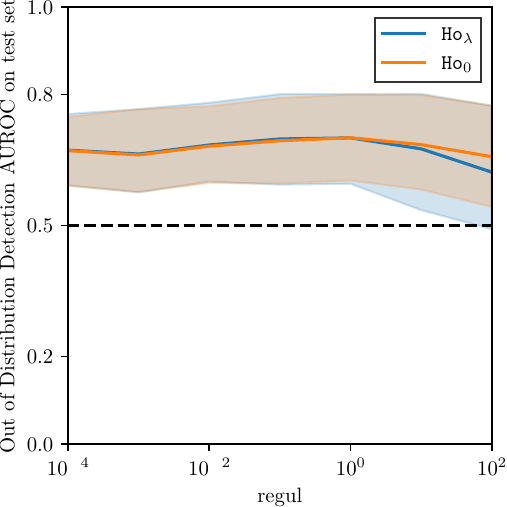}
    \includegraphics[width=0.6\linewidth]{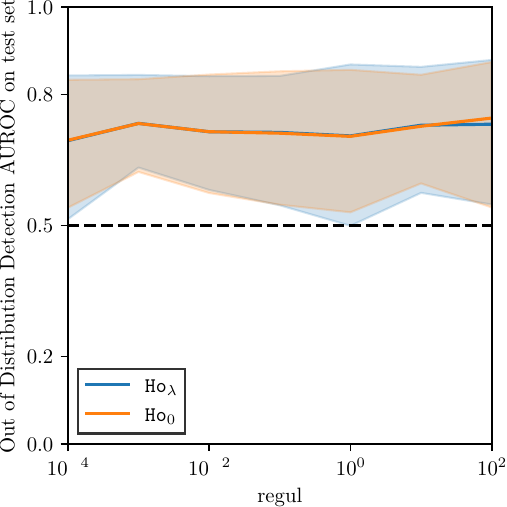}
    \caption{We compare the regularized $\texttt{Ho}_{\lambda}$ and the unregularized $\texttt{Ho}_0$ of the \texttt{Ho} uncertainty estimator on test examples of MNIST (top) and Fashion-MNIST (bottom) and measure whether examples with high uncertainty are examples from the missing subpopulation, for a number of $\lambda$ values.}
    \label{fig:ablation-lambda-sweep}
\end{figure}

\begin{figure}[H]
    \centering
    \includegraphics[width=0.6\linewidth]{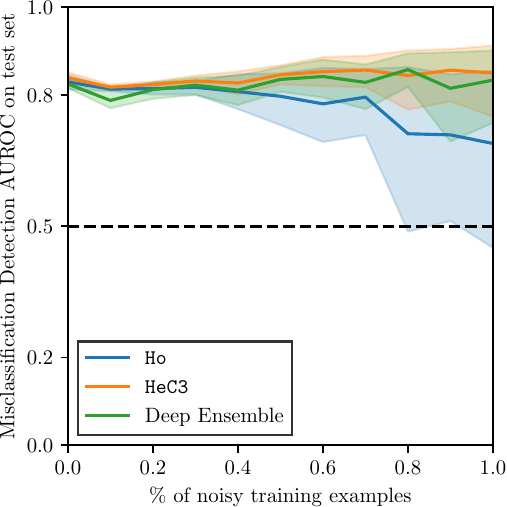}
    \includegraphics[width=0.6\linewidth]{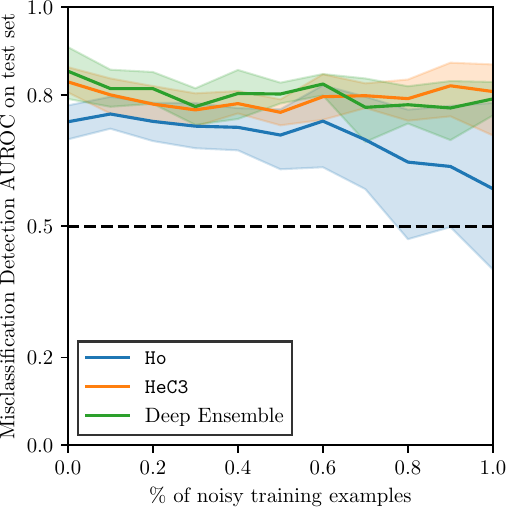}
    \caption{We compare different uncertainty estimators on test examples of MNIST (top) and Fashion-MNIST (bottom) and measure if examples with high uncertainty are mispredicted by the model, with varying level of label noise in the noisy subpopulation.}
    \label{fig:ablation-noise}
\end{figure}

\section{Qualitative Experiments}
\label{app:qual-xp}

\begin{table}[h]
\caption{Test examples of the IMDB review dataset sorted by their uncertainty. The upper part of the table corresponds to examples with relatively high \texttt{HeC3} uncertainty, and the bottom part to examples with high \texttt{Ho} uncertainty. 5 examples for each class are shown in the two columns of the dataframe.}
\begin{tabular}{lp{0.45\textwidth}p{0.45\textwidth}}
\toprule
 & \textbf{Negative review} & \textbf{Positive review} \\
\midrule
\multirow{24}{*}{\rotatebox[origin=c]{90}{\textbf{\texttt{HeC3} uncertain}}} & this movie isn't that great...at all but it's good when you want to just laugh, because it's pretty ridiculous :) there are a lot of mistakes in it and it's cheesy. i got this movie for Christmas like 5 years ago but for some reason i've never give... & quite good, don't expect anything high culture.......the acting is bad, the storyline fails, but it is still a fairly nice movie to watch. why? because it's dark, a little bit stupid, like unpredictable and just entertaining and fun to watch. do no...\\ & During my struggle to stay awake during this borefest, I fought through my near-dosing off to discover some silly plot regarding fraternity schmucks, quite incredibly obnoxiously annoying, running into trouble with a psychotic , radioactively damag... & Well, I'm an Italian horror big fan and I love movies from directors such Argento, Fulci, Bava Sr and Bava Jr, only to quote the most famous. "La villa delle anime maledette" is one of the most unknown movie of this genre, shot when this kind of ci...\\ & Just re-saw this last night and to put it bluntly: "Style instead of substance". We can already guess that there had to be a lot more to Jerry Lee Lewis than what is depicted here. The Jerry Lee Lewis character in this movie is not depicted as a re... & It is a damn good movie,with some surprising twists,a good cast and a great script. Only a couple of stupid bits,like the Rasta hit-man scene (This guy's a professional?) but that has been commented on already. The fact I had only heard one guy at ...\\ & I don't expect a lot from ghost stories, but I do expect a story to make a bit of sense! Is that asking too much from the screenwriters and filmmakers? When the bad guy, all of the sudden, becomes a homicidal maniac solely because a bunch of crows ... & Of course, how could he. He obviously co-opted several aspects from that excellent movie, which was also based on the sensational French case of the self-described "doctor in the World Health Organization" who murdered his family and himself when f...\\ & A woman as rich as she is insecure has a history of alcoholism and nervous breakdowns, helped no doubt by a smooth-talking gigolo husband who openly cheats on her. Naturally nobody believes her when she claims to have been accosted by a giant man w... & this movie is the best horror movie i have ever seen. the acting is terrible and the plot leaves a lot to be desired but the puppet gave me nightmares for weeks. seriously, if you have little kids don't let them see this. of course i am a little bi...\\ \midrule \multirow{24}{*}{\rotatebox[origin=c]{90}{\textbf{\texttt{Ho} uncertain}}} & (This might have a spoiler)When I first started watching this movie, I thought it was OK. The music was good and that bizarro dream sequence I was willing to forgive, but the lack of looks in the main character and all ... & Could this be one of the earliest colour films? It's actually the second. This is a very beautiful piece of film produced by Thomas Edison. This was one of many of his other films.I think this is the most beautiful of any ...\\ & Terry Benedict (Andy Garcia) catches up with Danny Ocean and his team and demands that they repay the money that they stole from him (in Oceans 11) plus interest. He holds back from violent action however as he is under the instruction of the world... & I knew next to nothing about this movie until I chanced to rent it. It was a very pleasant surprise. The cast is excellent including Matthau whom I do not normally care for. He makes a credible romantic lead. Hawn is a sweet kook and Bergman is tou...\\ & After the turning point of NIGHT MUST FALL, Robert Montgomery (for the most time) came into his finest films and performances: HERE COMES MR. JORDAN, THEY WERE EXPENDABLE, THE LADY IN THE LAKE, RIDE THE PINK HORSE, THE SAXON CHARM, JUNE BRIDE. Even... & This movie made by the NFBC was made in honor of the Montreal Canadians dynasty years in the 50's,60's and 70's. My 5th grade teacher played this in class in honor of my 11th birthday in 1987 and also to celebrate my return from a serious facial in...\\ & Supposedly, a movie about a magazine sending journalists to investigate reports of UFOs with one being more or less tolerant or agnostic about the whole affair and the other an Aussie, a hardened skeptic who laughs at the UFO nonsense. It's all a c... & By watching this film you will not only explore the "Turkish music" but will also explore the city of Istanbul with wonderful pictures and scenes from all over the important regions of the city.There are lots of delightful conversations with all so...\\ & I enjoy gay-themed movies where the characters aren't stereotypically gay and that's what attracted me to this movie, that and the principal actor, which is the only reason I'm giving this movie one more star than it deserves (although not because ... & I saw Riverdance - The New Show and loved it from the very first moment! It is an energetic tribute to Irish dance filled with brilliant dancing, music and choreography! The leads, Jean Butler and Colin Dunne had me captivated with their exquisite ...\\
\bottomrule
\end{tabular}
\end{table}

\begin{figure}[H]
\centering
\begin{tabular}{cc}
  \includegraphics[height=0.2\linewidth]{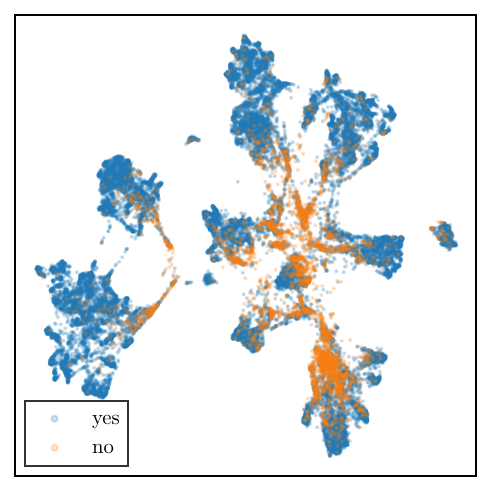}& 
\includegraphics[height=0.2\linewidth]{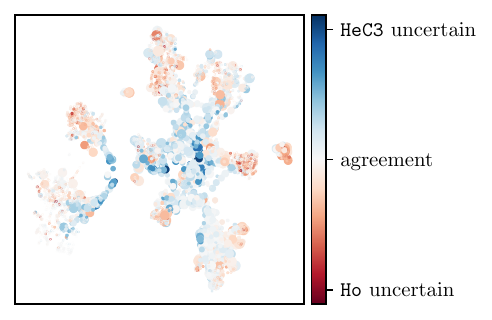}
\includegraphics[height=0.2\linewidth]{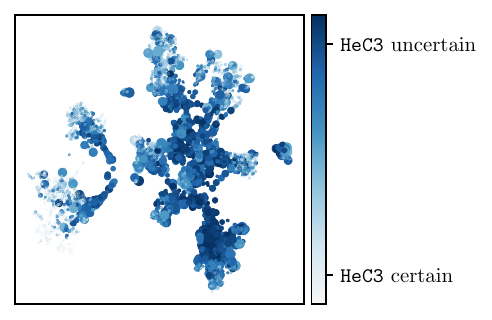}\\

\includegraphics[height=0.2\linewidth]{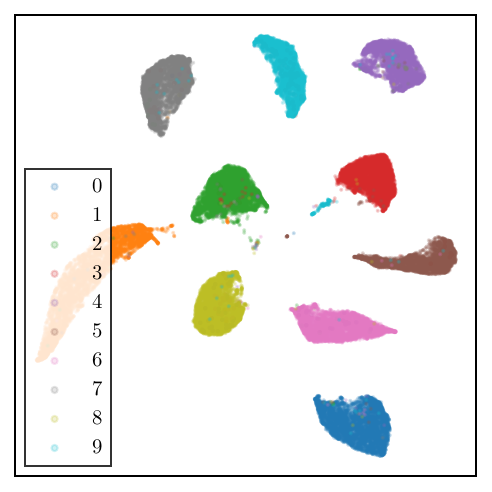}& 
\includegraphics[height=0.2\linewidth]{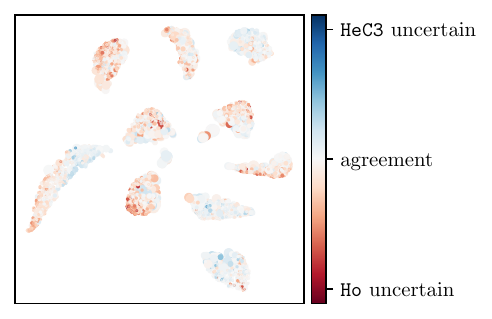}
\includegraphics[height=0.2\linewidth]{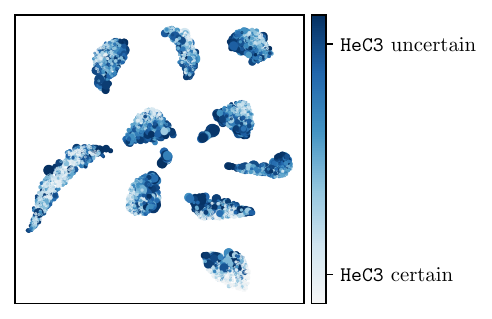}\\

  \includegraphics[height=0.2\linewidth]{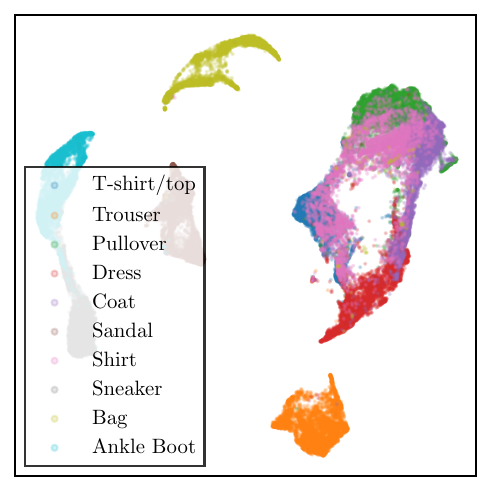}&   \includegraphics[height=0.2\linewidth]{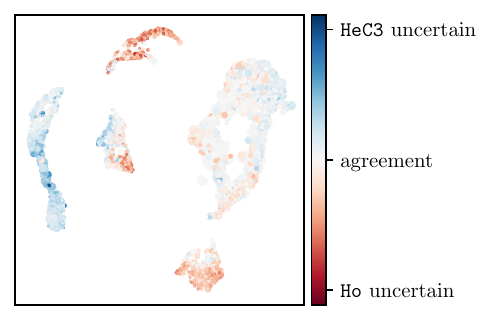} \includegraphics[height=0.2\linewidth]{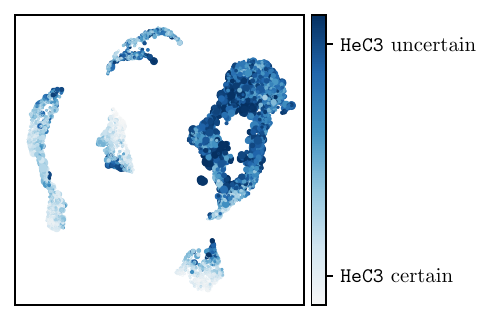}\\
  \includegraphics[height=0.2\linewidth]{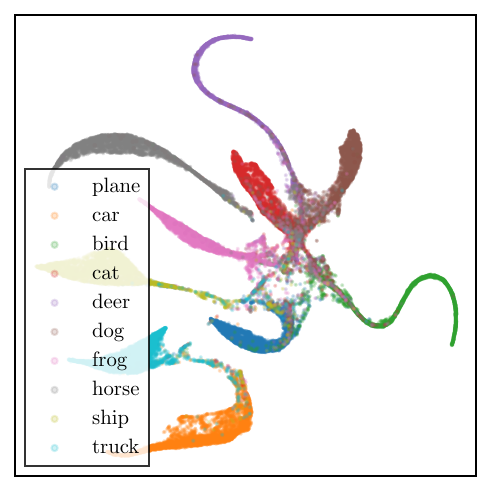}&   \includegraphics[height=0.2\linewidth]{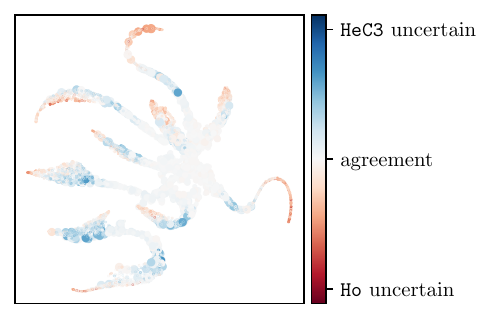}  \includegraphics[height=0.2\linewidth]{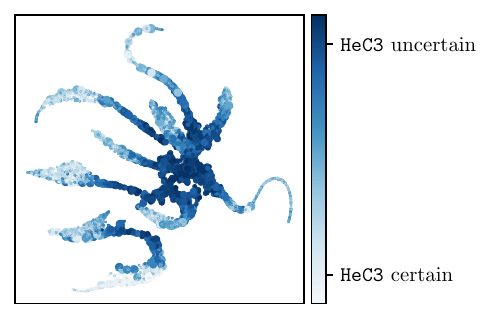}\\
  
\end{tabular}
    \caption{umap \citep{mcinnes2018umap} projections of last layer embeddings of the different neural network architectures trained on (from top to bottom) \emph{bank-marketing} (MLP), MNIST (MLP), Fashion-MNIST (ConvNet), and CIFAR-10 (ResNet18). In the left column, train samples and their classes are highlighted in color. In the middle column, we compute the agreement or disagreement between \texttt{Ho} and \texttt{HeC3}. In the right column, we the gradient of color is with respec to \texttt{HeC3} only.}
\end{figure}

\begin{figure}[H]
\centering
\begin{tabular}{cc}
  \includegraphics[width=0.35\linewidth]{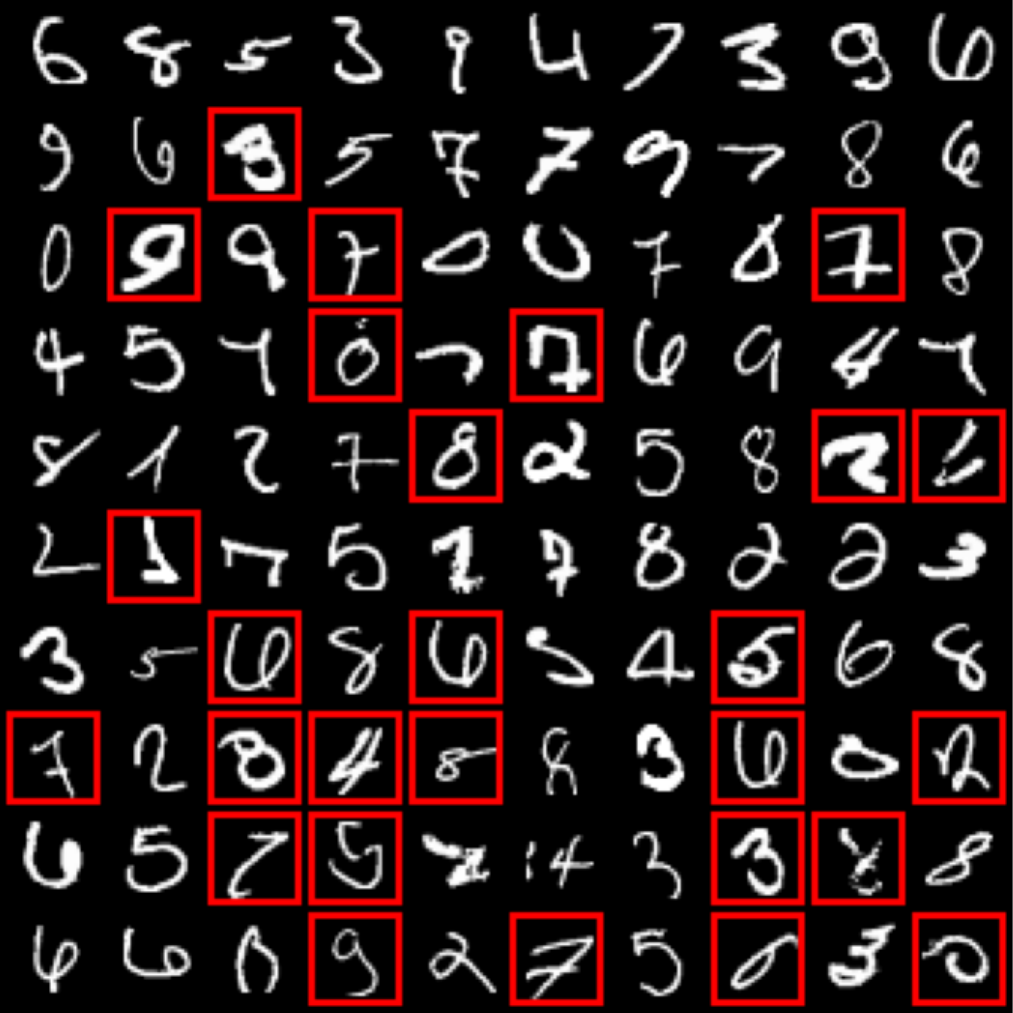}&   \includegraphics[width=0.35\linewidth]{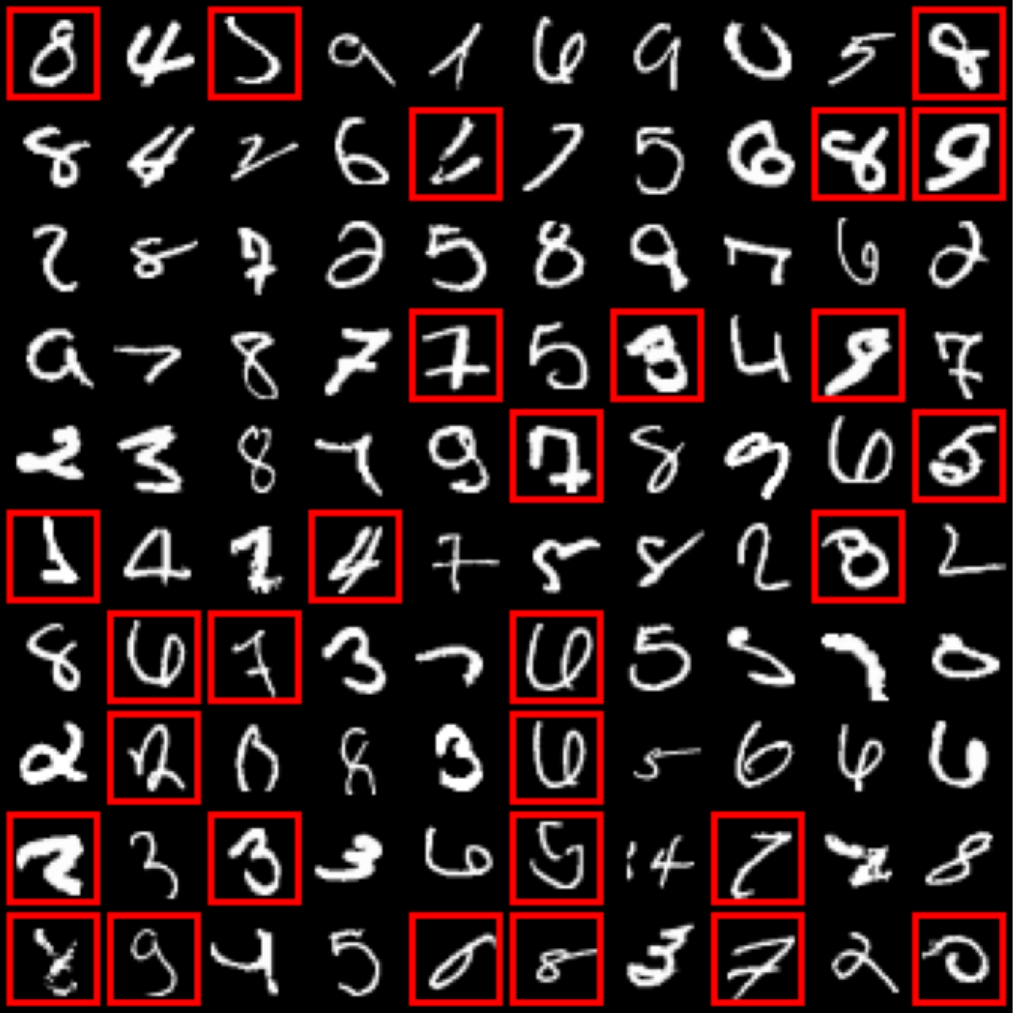}\\
  \includegraphics[width=0.35\linewidth]{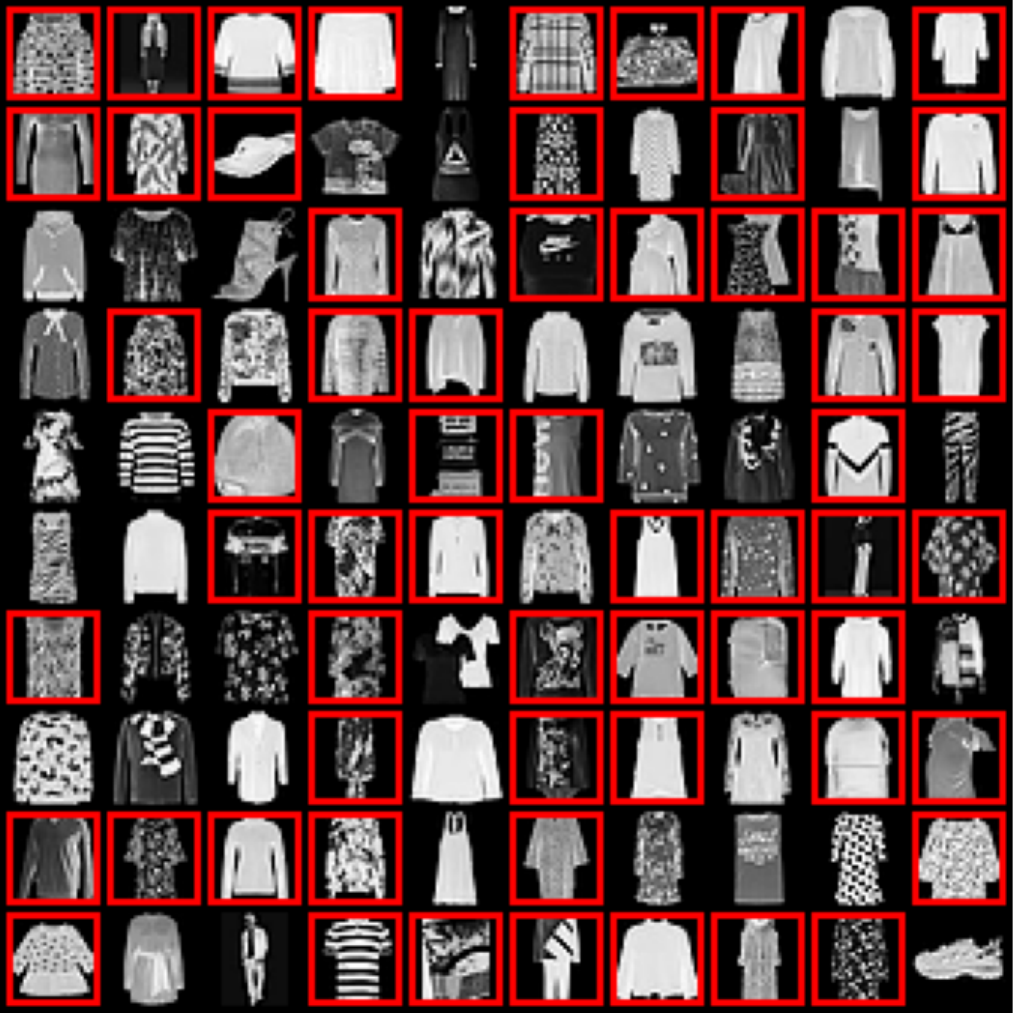}&   \includegraphics[width=0.35\linewidth]{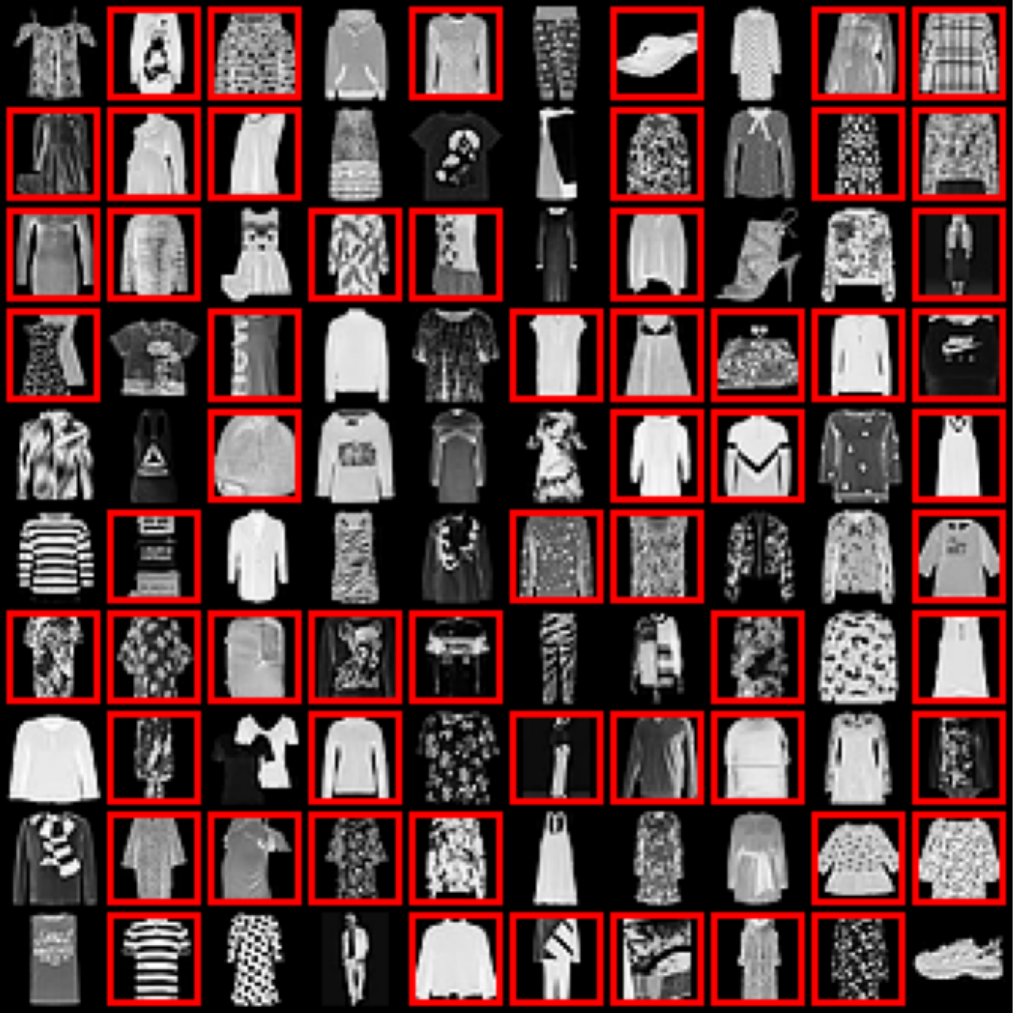}\\
  \includegraphics[width=0.35\linewidth]{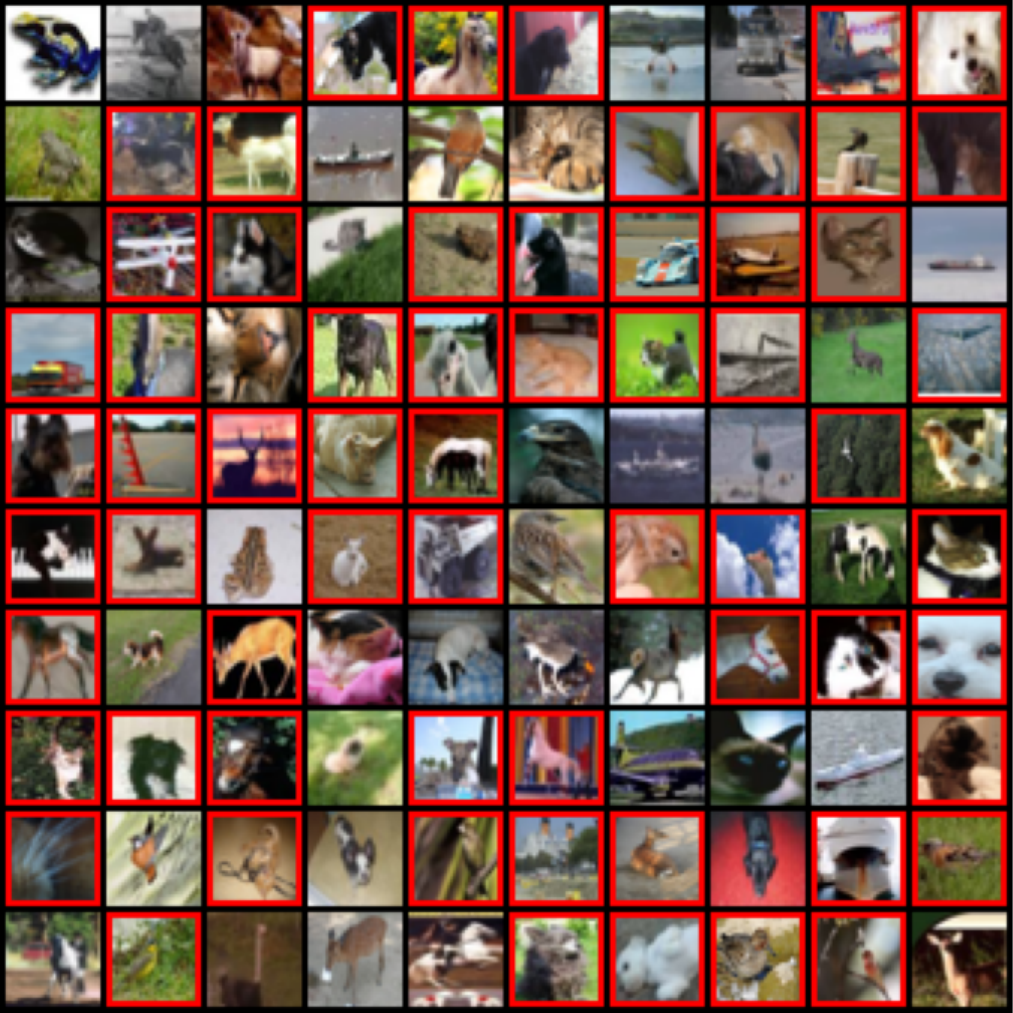}&   \includegraphics[width=0.35\linewidth]{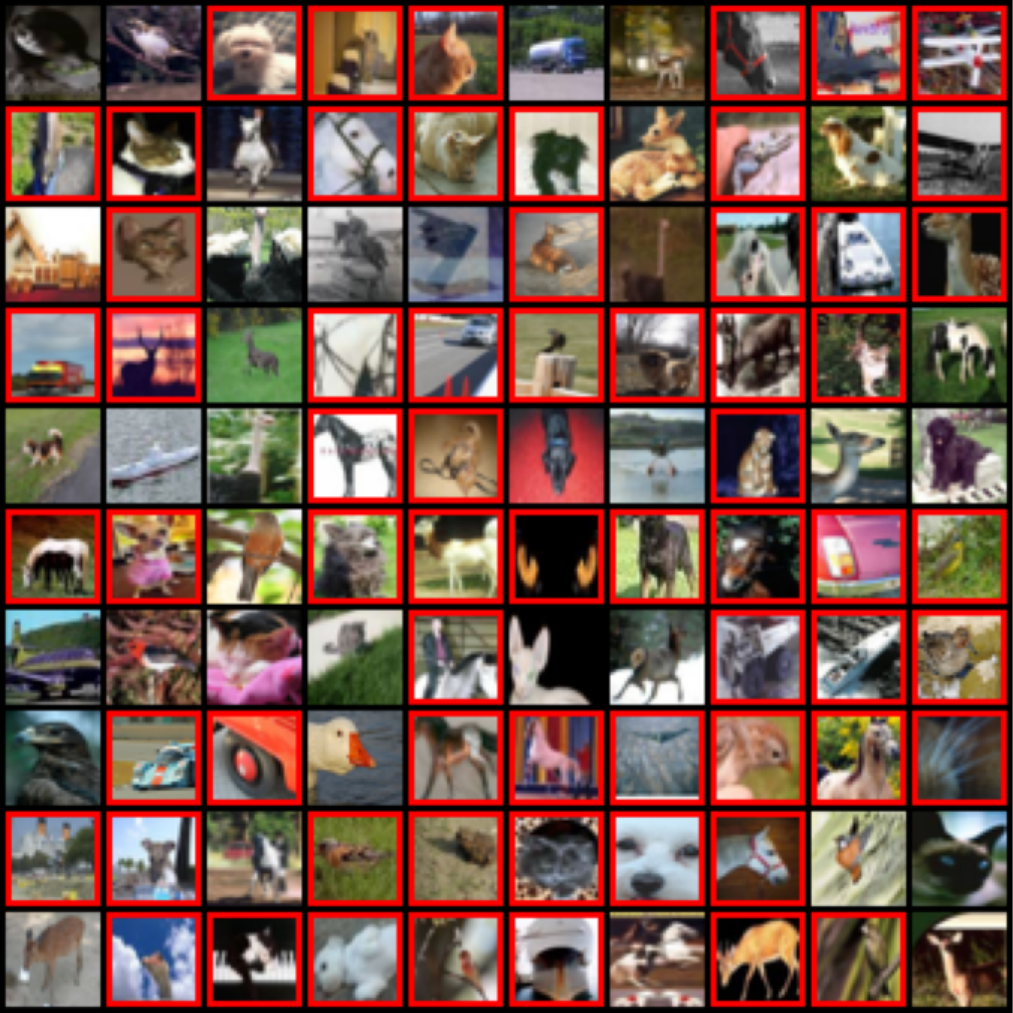}\\
\end{tabular}
\caption{Test examples of MNIST, Fashion-MNIST and CIFAR-10 sorted by their variances. The 100 examples with the highest jackknife variance are grouped in the left grid, and the 100 examples with the highest MLE variance are grouped in the right grid. Examples wrongly predicted in the test set are highlighted in red.}\label{fig:toptop}
\end{figure}

\begin{figure}
\centering
\begin{tabular}{cc}
  \includegraphics[width=0.35\linewidth]{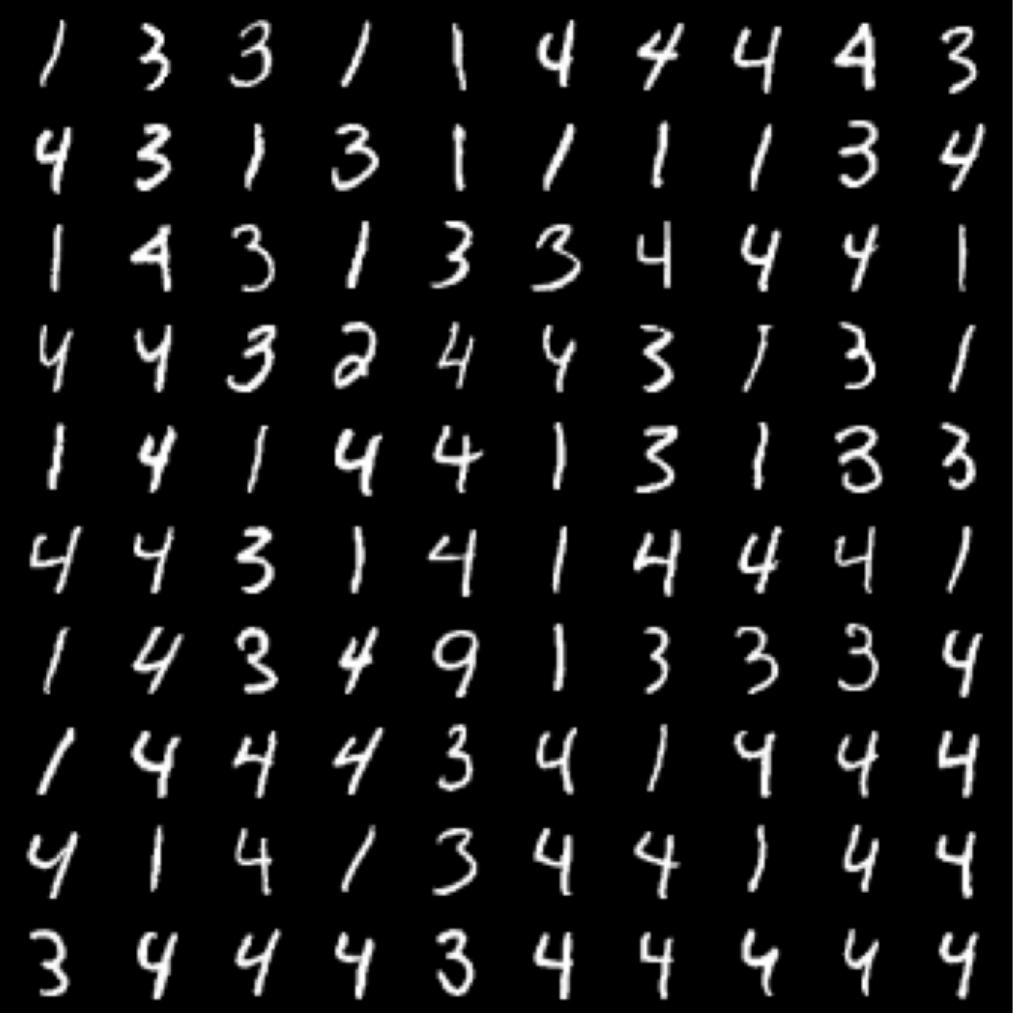}&   \includegraphics[width=0.35\linewidth]{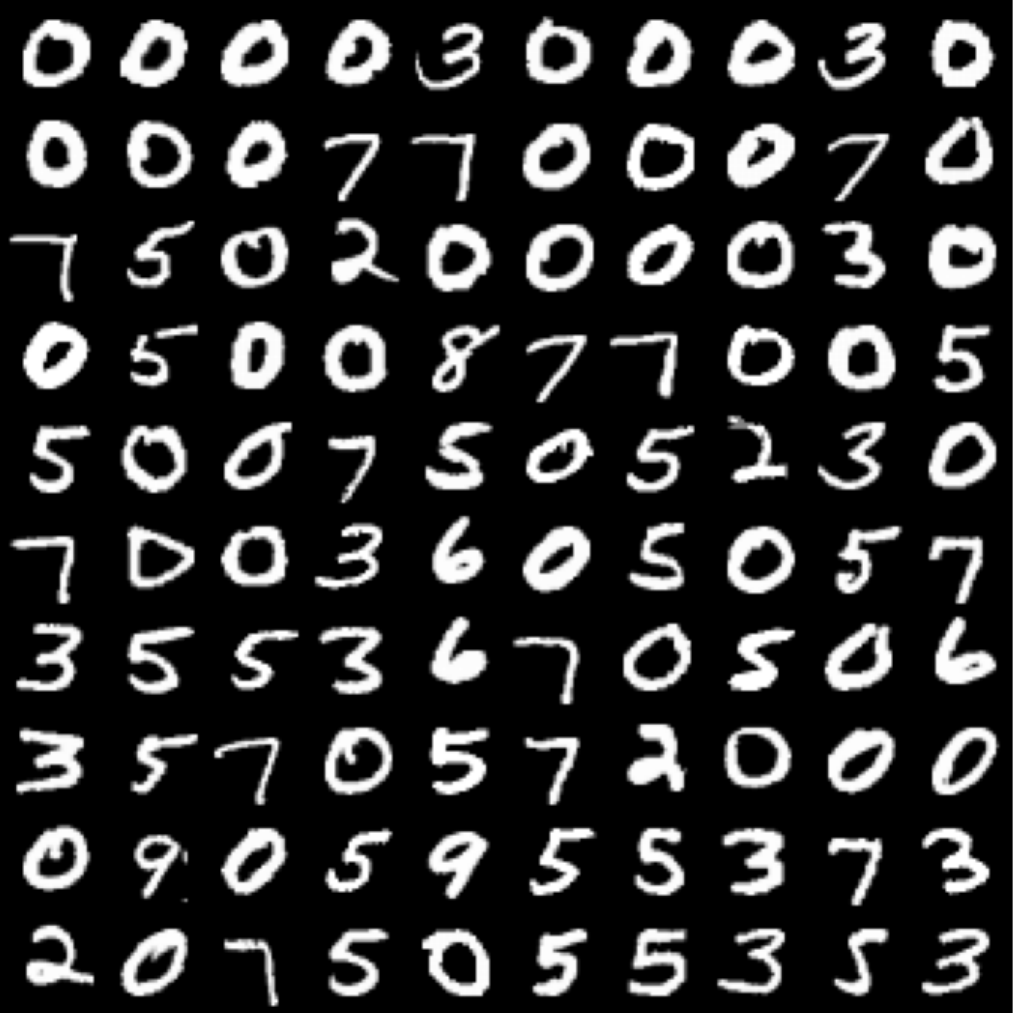}\\
  \includegraphics[width=0.35\linewidth]{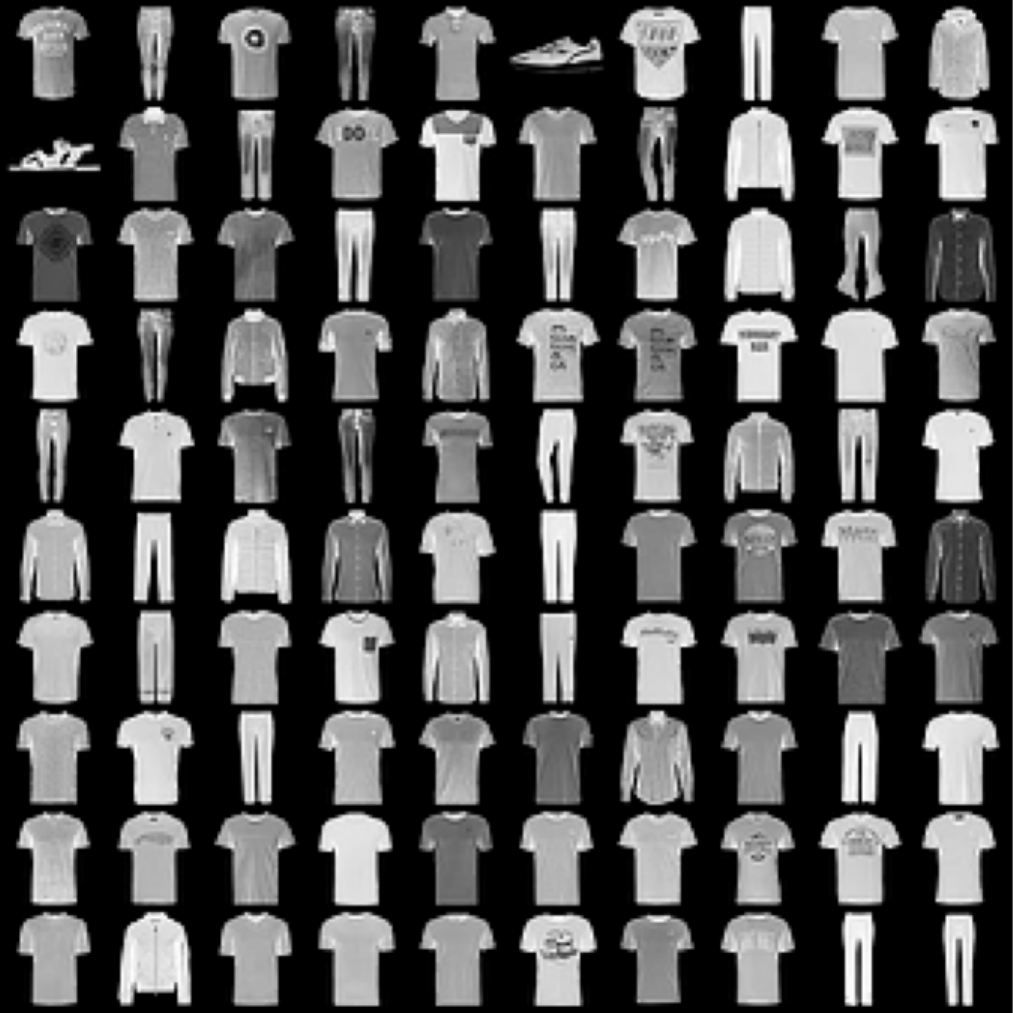}&   \includegraphics[width=0.35\linewidth]{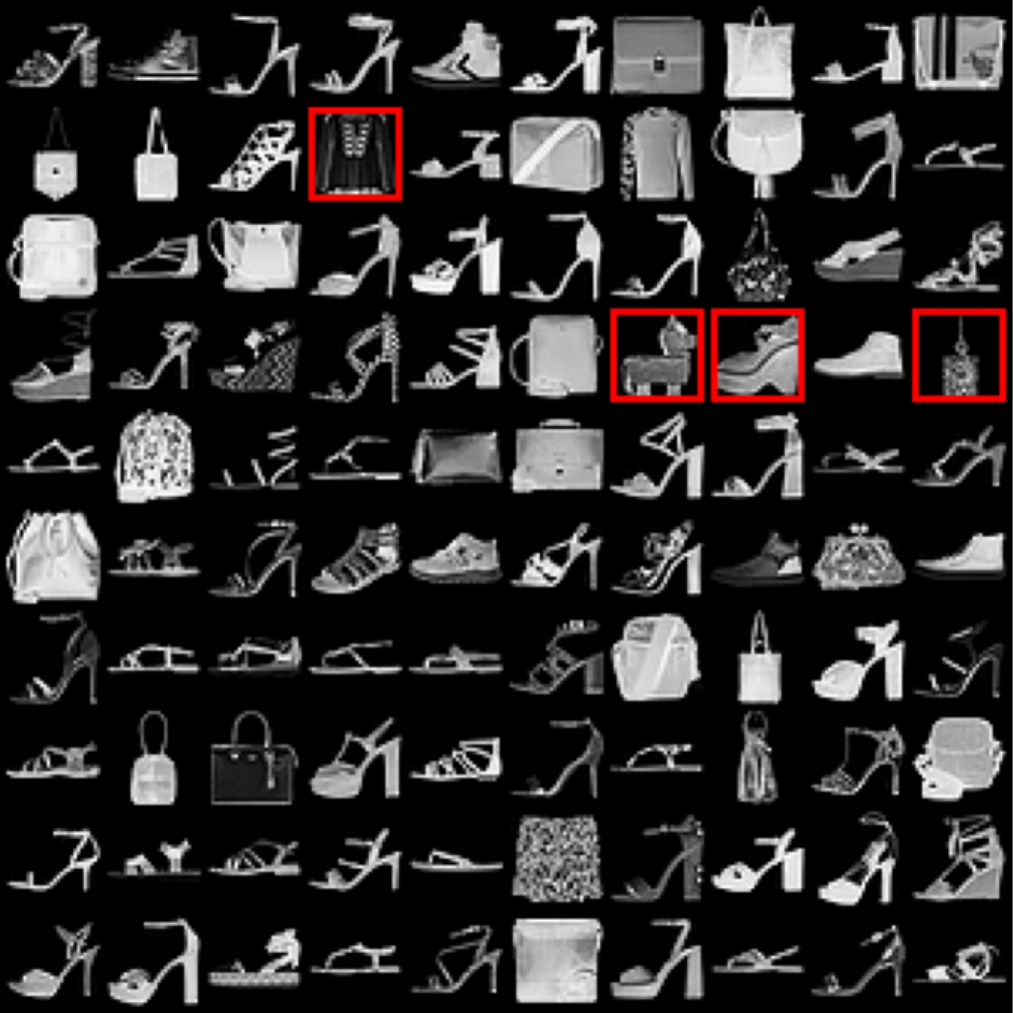}\\
  \includegraphics[width=0.35\linewidth]{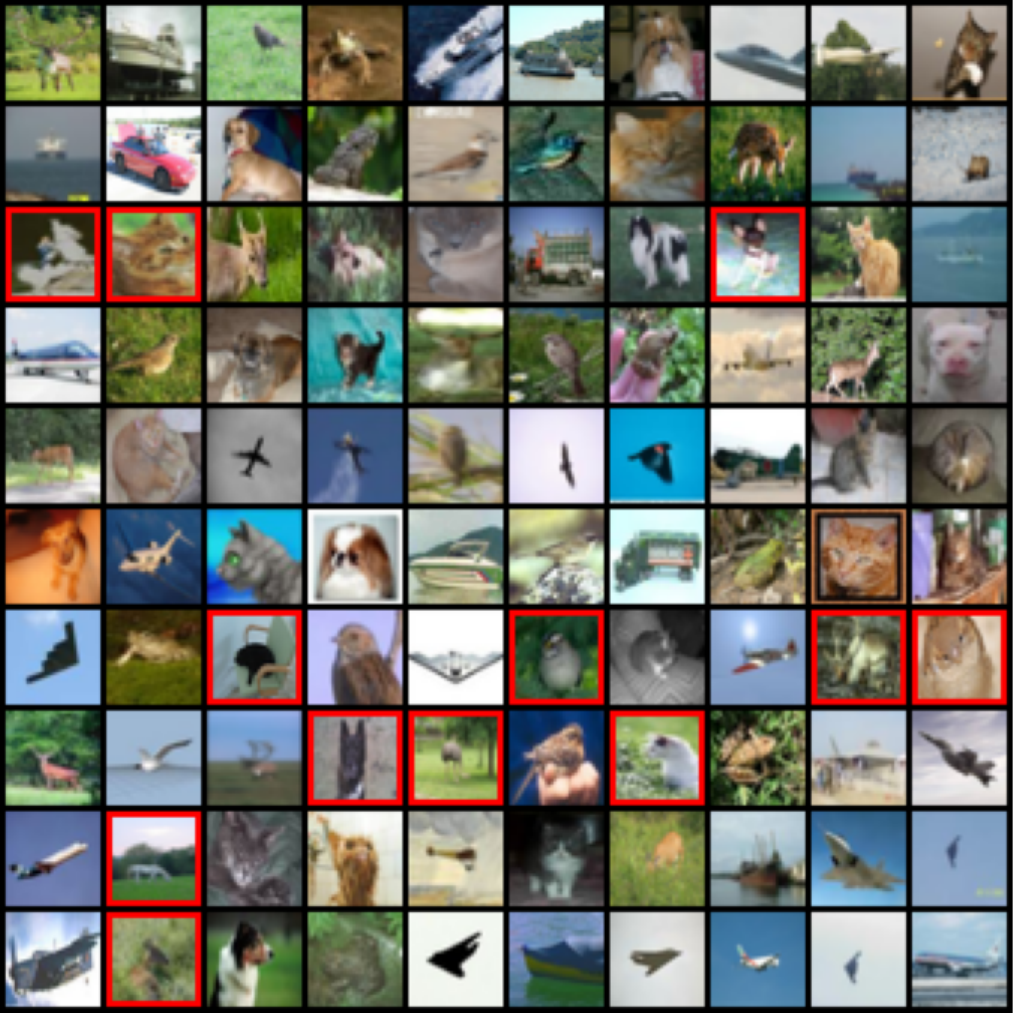}&   \includegraphics[width=0.35\linewidth]{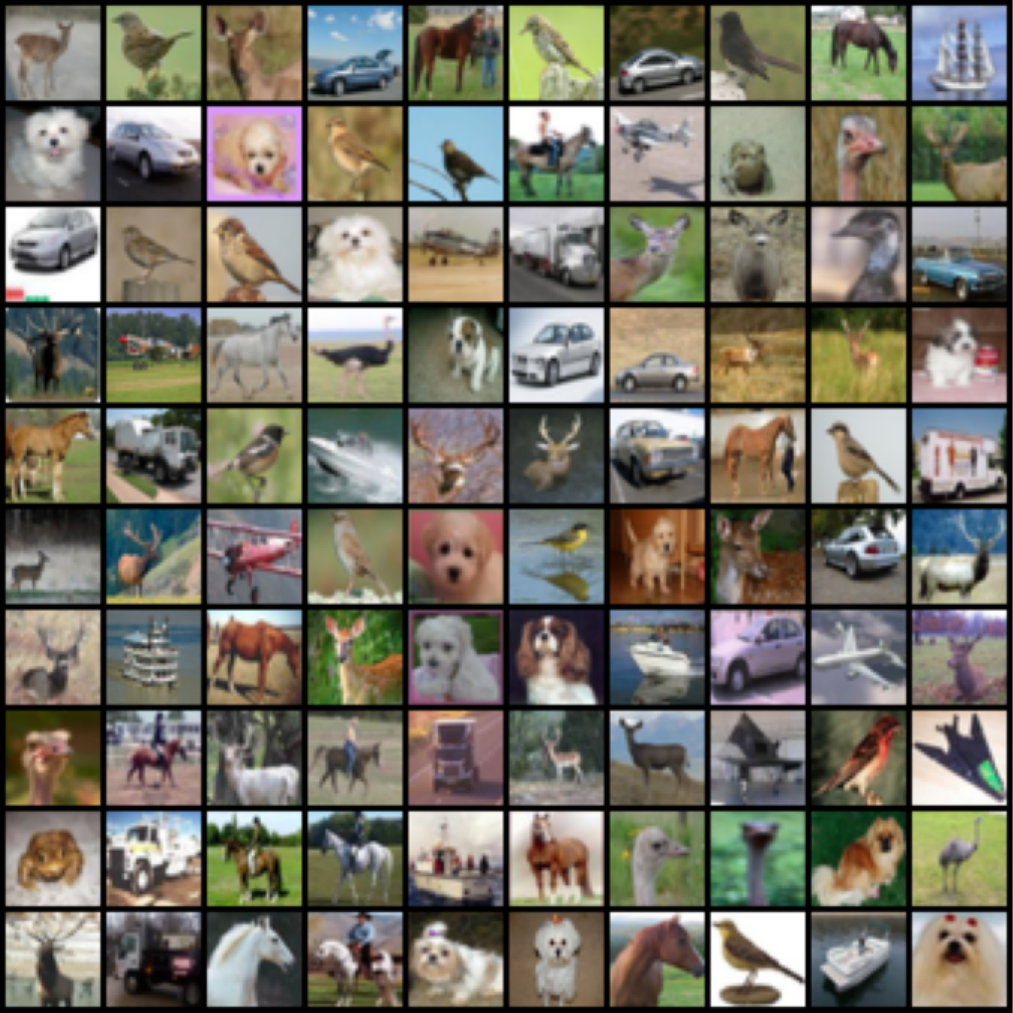}\\
\end{tabular}
\caption{Test examples of MNIST, Fashion-MNIST and CIFAR-10 sorted by the agreements between their $\var_{\text{JK}}$ and $\var_{\text{MLE}}$ variances. The 100 examples with $\var_{\text{JK}}>\var_{\text{MLE}}$ are grouped in the left grid, and the 100 examples with $\var_{\text{MLE}}>\var_{\text{JK}}$ are grouped in the right grid. Examples wrongly predicted in the test set are highlighted in red.}
\end{figure}

\end{document}